\pdfoutput=1
\documentclass[11pt]{article}
\usepackage[table]{xcolor}
\usepackage{acl}

\usepackage{times}
\usepackage{latexsym}

\usepackage[T1]{fontenc}

\usepackage[utf8]{inputenc}

\usepackage{microtype}

\usepackage{multirow}
\usepackage{CJKutf8}

\usepackage{kotex}
\usepackage{tabularx}
\usepackage{graphicx}
\graphicspath{ ./2022figures/ } 
\usepackage{color, xcolor}
\usepackage{colortbl,booktabs}

\usepackage{soul}
\soulregister\cite7 
\soulregister\citep7 
\soulregister\citet7 
\soulregister\ref7
\soulregister\pageref7 
\usepackage{hyperref}

\title{\textsc{mDIA}: A Benchmark for Multilingual Dialogue Generation \\ in 46 Languages}

\author{Qingyu Zhang$^{1}$\thanks{$\>$ Authors contributed equally.}, Xiaoyu Shen$^{2}$\footnotemark[1], Ernie Chang$^{2}$, Jidong Ge$^{1}$\thanks{$\>$ Corresponding author. Email to gjd@nju.edu.cn} and Pengke Chen$^{1}$ \\
$^1$State Key Laboratory for Novel Software Technology,  Nanjing University\\
$^2$Department of Language Science and Technology, Saarland University
}

\date{}

\begin{document}
\maketitle
\begin{abstract}
Owing to the lack of corpora for low-resource languages, current works on dialogue generation have mainly focused on English.
In this paper, we present \textsc{mDIA}, the first large-scale multilingual benchmark for dialogue generation across low- to high-resource languages. It covers real-life conversations in 46 languages across 19 language families. We present baseline results obtained by fine-tuning the multilingual, non-dialogue-focused pre-trained model mT5 as well as English-centric, dialogue-focused pre-trained chatbot DialoGPT. The results show that mT5-based models perform better on sacreBLEU and BertScore but worse on diversity. Even though promising results are found in few-shot and zero-shot scenarios, there is a large gap between the generation quality in English and other languages.
We hope that the release of \textsc{mDIA} \footnote{The dataset and processing scripts are available in \url{https://github.com/DoctorDream/mDIA}.}  could encourage more works on multilingual dialogue generation to promote language diversity.

\end{abstract}

\section{Introduction}

Human-machine conversation has been a critical and challenging task in AI and NLP~\cite{weizenbaum1966eliza,walker2001quantitative}. In the past few years, rapid progress has been made in open-domain dialogue generation. By training an end-to-end seq2seq model~\cite{sutskever2014sequence} on billions of conversational data, it is able to generate human-like responses and even outperform commercial systems built on complex pipelined frameworks~\cite{zhang2020dialogpt,su2020moviechats,adiwardana2020towards,roller2021recipes}. Nonetheless, large amounts of conversational data is only available in English and a few number of high-resource languages~\cite{razumovskaia2021crossing}. For most of the 6,500 languages in the world, there are not even any public datasets for research. Their development in dialogue generation has largely lagged behind~\cite{joshi2020state,blasi2021systematic}.

Current multilingual conversational corpora are mainly for task-oriented dialogues and are constructed by translating English data ~\cite{lin2020xpersona,zhu2020crosswoz,ding2021globalwoz}, which is not suitable for open-domain chats. \cite{sato2018addressee} collect a dataset for response selection in 12 languages, but it covers only technical discussions in Ubuntu IRC logs, which is far from everyday conversations.
\cite{zhang-etal-2021-dataset} present a multilingual reply suggestion dataset with ten languages, but it pay more attention to construct the dataset and benchmark on higher-resource languages.
A few conversational datasets for low-resource languages exist~\cite{kowsher2019doly,alam2018neural,naous2020empathy,adewumi2021sm}, but they are created in different domains and studied independently. A unified benchmark to track and evaluate multilingual dialogue generation across low- to high-resource languages is still lacking.

\begin{table}[!t]
\small
\centering	
{
\begin{tabularx}{0.5\textwidth}{X X}

\rowcolor[rgb]{ .878,  .878,  .878} Korean \\
\textcolor[rgb]{ .651,  .651,  .651}{[Context]}
\textcolor[rgb]{ .325,  .553,  .835}{플로리다는 화씨 75도예요. 너무 따뜻해요.} \\
\textcolor[rgb]{ .651,  .651,  .651}{[Response]}
\textcolor[rgb]{ .325,  .553,  .835}{화씨 75도면 섭씨로 23.3도인가요? 서울로 치면 가을 날씨네요! 여름에도 온도가 비슷해요?} \\
\rowcolor[rgb]{ .878,  .878,  .878} Japanese \\
\textcolor[rgb]{ .651,  .651,  .651}{[Context]}
\textcolor[rgb]{ .886,  .42,  .039}{できた後の満面の笑顔がかわいい   画面の下でワンコも喜んでるね} \\
\textcolor[rgb]{ .651,  .651,  .651}{[Response]}
\textcolor[rgb]{ .886,  .42,  .039}{最高の瞬間！笑顔にしてくれる} \\
\rowcolor[rgb]{ .878,  .878,  .878} English \\
\textcolor[rgb]{ .651,  .651,  .651}{[Context]}
\textcolor[rgb]{ .463,  .576,  .235}{What is the meaning of life?} \\
\textcolor[rgb]{ .651,  .651,  .651}{[Response]}
\textcolor[rgb]{ .463,  .576,  .235}{There isn’t one. So make your own.} \\

\end{tabularx}

}
\caption{\label{tab:example}\small Examples of multilingual conversations in \textsc{mDIA}.} \vspace{-3mm}
\end{table}

To fill this gap, we present \textsc{mDIA}: a large-scale multilingual dialogue generation benchmark. It contains real-life conversations in 46 languages crawled from the full-year Reddit\footnote{\url{https://www.reddit.com/}} traffic in 2020. As all the data originates from the same source, they tend to exhibit similar styles and topics, making \textsc{mDIA} a suitable benchmark to track, improve and compare multilingual dialogue generation holistically. Table~\ref{tab:example} shows some conversation examples in \textsc{mDIA}. With this benchmark, our main focus is to explore the following research question: ``\emph{Given limited training data, how can we make the most of current techniques to improve dialogue generation in a low-resource language?}''. We restrict the number of conversations to be at most 12k for every single language since we focus on the low-resource scenario. Indeed, in the final \textsc{mDIA} benchmark, only half of the languages have 12k conversational exchanges available. The rest of languages have a range of $165\sim 9525$ conversational exchanges, which further confirms the necessity of exploring multilingual dialogue generation techniques in the low-resource setting.

We explore two different directions to tackle this problem. The first leverages mT5~\cite{xue2021mt5}, a SOTA multilingual language model pre-trained on 101 languages. mT5 is equipped with \emph{language-specific knowledge but is not tailored for dialogue generation}. The second leverages DialoGPT~\cite{zhang2020dialogpt}, a well-developed English chatbot trained on 147M conversation-like exchanges. DialoGPT is \emph{tailored for dialogue generation but is focused only on English}.


We find that mT5-based models usually perform better for relevance but worse for diversity. Mixing all languages and training a single model helps low-resource languages but affects marginally for high-resource ones. A single multilingual mT5-based model also shows promising results in generalizing to unseen languages and is able to produce valid responses even in mixed languages. Nonetheless, the generation quality for all other languages is far behind that of English, suggesting significant research is required for cross-lingual transfer of conversing skills.

To summarize, we make the following contributions: (1) We present \textsc{mDIA}, the first multilingual dialogue generation benchmark covering 46 languages across 19 language families. (2) We release tools for extracting multilingual conversations from the Reddit traffic, with which can further enlarge the dataset. (3) We provide strong baseline results and findings to benefit future research.

\section{Related Works}
Building intelligent open-domain chatbots that can converse with humans coherently has been a long-standing goal of artificial intelligence~\cite{huang2020challenges}. Unlike task-oriented dialogue systems targeting at one specific goal and domain, open-domain chatbots do not have any restriction on the conversation topic. This makes them much more challenging to develop and evaluate. Earlier chatbot systems are built upon complex pipelines with rule-based templates and can only work in a constrained environment~\cite{weizenbaum1966eliza,colby1971artificial,wallace2009anatomy}. Recently, with the rapid advances in large-scale pre-trained language models~\cite{brown2020language,lewis2020bart}, open-domain chatbots have been able to produce human-like responses by finetuning on billions of conversational data~\cite{zhang2020dialogpt,adiwardana2020towards,roller2021recipes}. Nonetheless, current works focus almost exclusively on English. As pre-trained models are extremely data hungry, repeating the training process as done in English is only feasible for very few high-resource languages like Chinese~\cite{su2020moviechats,li2020empirical} and Japanese~\cite{sugiyama2021empirical}. 

There have been small-scale conversational datasets for low-resource languages like Urdu~\cite{alam2018neural}, Arabic~\cite{naous2020empathy}, Swedish~\cite{adewumi2021sm} and Bengali~\cite{kowsher2019doly}. They are constructed from different sources and are studied independently. Some multilingual conversational corpora exist for task-oriented dialogues in one specific domain by translating the English data~\cite{lin2020xpersona,zhu2020crosswoz,ding2021globalwoz}, which is far from informal open-domain chats as every language has its own chatting style different from English~\cite{moorjani1988semiotic}. \cite{sato2018addressee} cover only technical discussions from the Ubuntu IRC logs and are very different from everyday conversations. \cite{zhang-etal-2021-dataset} cover open-domain dialogue data with ten languages that focus on high-resource languages.

Quite a few multilingual datasets exist for tasks other than dialogue generation, most of which are focused on classification tasks~\cite{lewis2020mlqa,hu2020xtreme,liang2020xglue,lin2020xpersona}. Many cross-lingual transfer techniques have been proposed for them and promising results can be achieved by finetuning only on English labeled data~\cite{ruder2021xtreme}. For generation tasks, however, cross-lingual transfer is much more challenging due to the necessity of language alignment in both the encoder and decoder side~\cite{vzagar2021cross,bai2021cross}. Directly finetuning the multilingual pre-trained language model or using a pivot-based approach remain strong baselines for many generation tasks~\cite{montero2020pivot,kale2020machine,fan2020multilingual,hasan2021xl,chen2021mtg,adelani2022thousand}.

\section{Dataset Construction}
\begin{figure}[t!]
 
\centering
\includegraphics[width=0.48\textwidth]{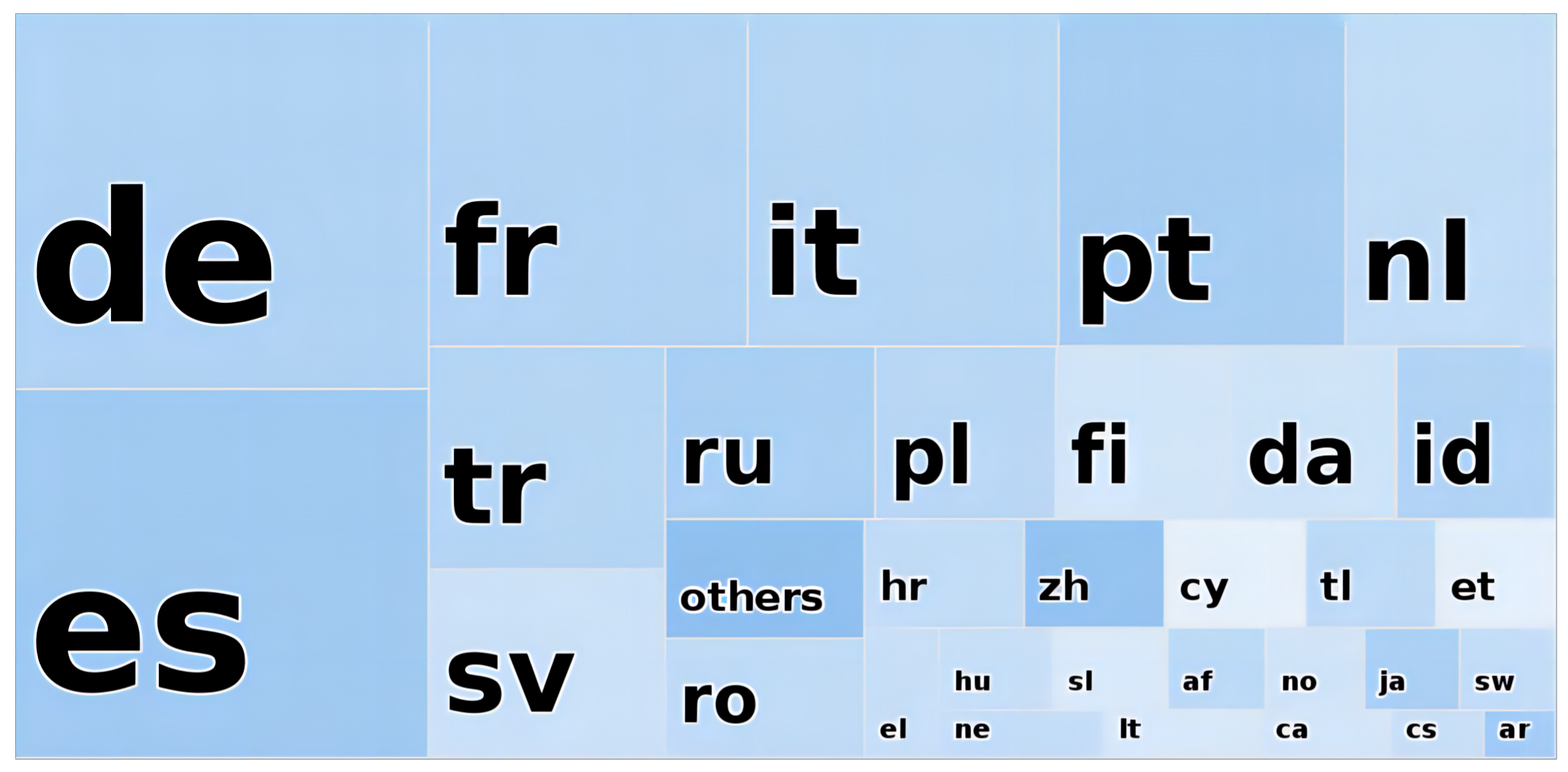}        
\caption{\small Distribution of non-English comments in Reddit 2020. Darker colors indicate more native speakers.}
\label{fig:distribution}
 
\end{figure}

\noindent\textbf{Data Source} We collect multilingual conversations from Reddit, one of the largest online discussion forum. However, Reddit has no strict classification of languages making it difficult to directly extract dialogues with low-resource languages. Although Reddit exists language-specific subreddits but they are rarely active for low-resource langauges. From an analysis of the whole 200M comments from January 2020, we find large amounts of low-resource languages concentrate on popular subreddits like ``r/askreddit", ``r/funny". Therefore, we take an approach whereby we first download all user comments from the whole year of 2020\footnote{\url{https://files.pushshift.io/reddit/comments}} using praw,\footnote{\url{https://github.com/praw-dev/praw}} then extract conversations in different languages from them. The full year traffic is 2.5TB consisting of 3 billion comments, out of which 95\% are in English. Even among non-English languages, the distribution is highly unbalanced. As shown in Figure~\ref{fig:distribution}, the top 5 languages constitute over half of all comments. It also correlates poorly with the number of native speakers. Many languages like Chinese, Japanese and Arabic, despite having huge numbers of native speakers, are rather under-represented in Reddit.

\noindent\textbf{Dialogue Extraction} 
After downloading the full-year traffic data we call the Reddit API to map each comment to their parent comment. Each parent comment and the comment itself form a context-response pair. We first apply \textit{langid}\footnote{\url{https://github.com/saffsd/langid.py}} 
with threshold $0.8$ to classify the language of each comment. However, we find quite a few misclassified languages due to the interference of emojis even after we filter most emojis defined by rules on emojipedia~\footnote{\url{https://emojipedia.org/}}. 
As the list of emojis is an open set and some unicodes of emojis overlap with normal characters, fully removing all emojis is difficult. Therefore, we apply a second-stage filter with \textit{langdetect}~,\footnote{\url{https://github.com/Mimino666/langdetect}} which is more robust with emojis. A dialogue is identified as in one language if both the context and response are in that language classified by both language detection tools \textit{langid} and \textit{langdetect}. 
Our human evaluation suggests that The construction form of context-response pairs is an effective form of dialogue, and it is beneficial to obtain more multilingual dialogue data. We will expand the single-round dialogue to the form of multiple rounds in the follow-up work as well.

\begin{figure}[t!]
 
\centering
\includegraphics[width=0.5\textwidth]{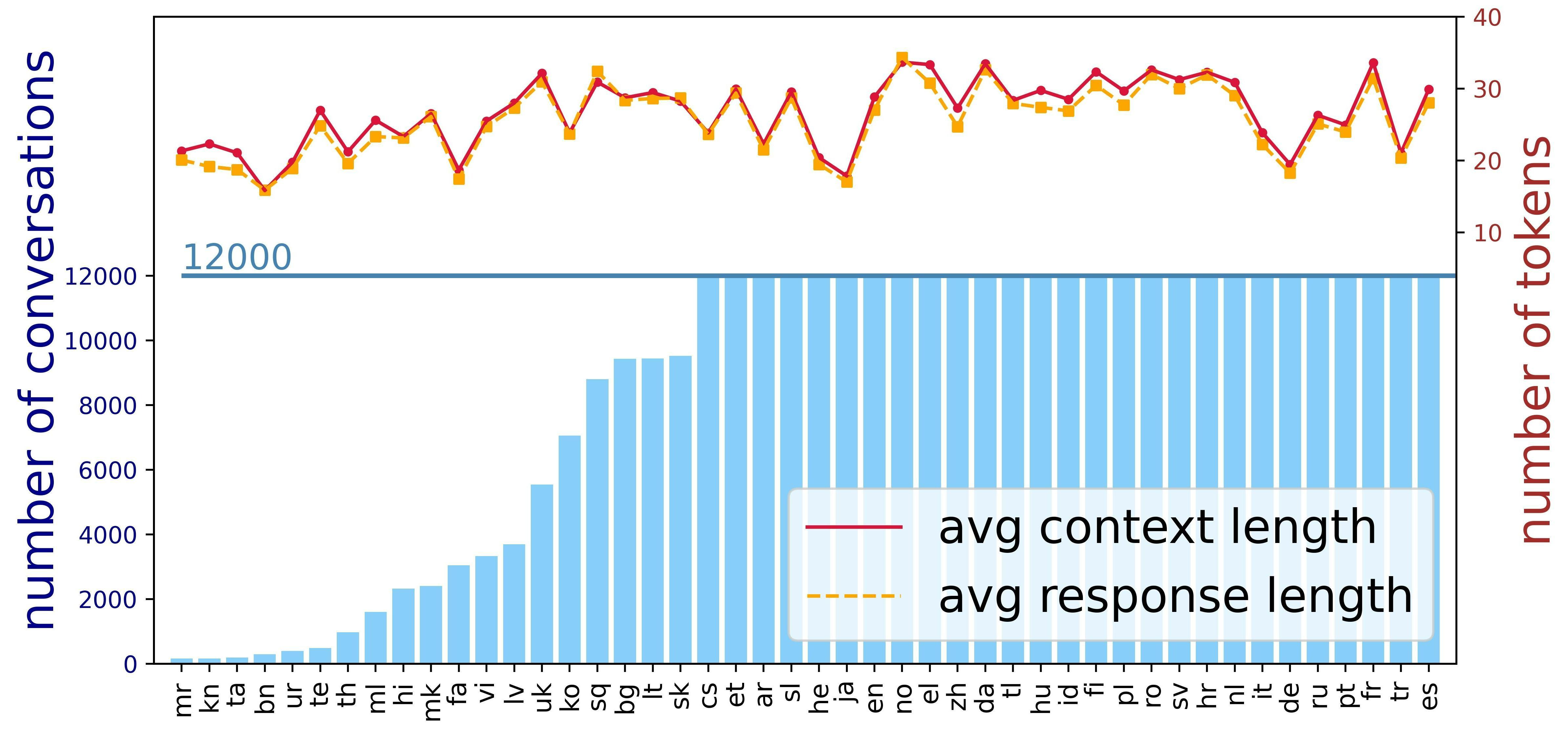}        
\caption{\small Statistics of \textsc{mDIA}. 26 non-English languages have 12k conversations and the remaining 19 have less than 12k available.}
\label{fig:statistics}
 
\end{figure}

\noindent\textbf{Dialogue Filtering} For all context-response pairs identified as same language, we apply several filters to reduce noise: (1)~remove blank contents, e.g., comments with the [delete] tag only, (2)~remove comments with over 2 line breaks to reduce unusual conversations like poems and long remarks, (3)~replace all URLs and user names with [URL] and [USER] tags, (4)~remove it if the length of the context or response is not within [5,79], to avoid generic or over-complicated contents,\footnote{We count the length based on the mT5 tokenizer.}, (5)~remove repeated comments to ensure all context and response in our datasets are unique and (6)~remove toxic comments by filter rules referred in Section~\ref{sec:ethics}. After filtering, we keep at most 12k context-response pairs for every language. Note that there can be certainly more conversational data from Reddit or other country-specific forums. The focus of this paper is \textbf{NOT} to achieve SOTA performance for each individual language given all available data, but rather to measure how we can transfer dialogue generation techniques to an arbitrary language in the low-resource setting, so we keep 12k as an upper limit. Future work can improve the performance on individual languages with more training data.

\noindent\textbf{Statistics} After filtering and language detection, We identified 45 languages (excluding English) from all Reddit traffic in 2020. We show the final statistics of \textsc{mDIA} in Figure~\ref{fig:statistics}. Half of them have 12k conversations available and the other half do not. 7 languages have fewer than 1,000 conversations for each. The lowest-resource languages Marathi and Kannada have only 165 and 167 conversations available. For these 7 languages, we keep all as the testset. For the others, we keep 1,000 conversations as the test set and split the rest with 10:1 for training/validation. We also show the average length of the context and response in Figure~\ref{fig:statistics}. Almost all languages have an average length of $15\sim 30$ tokens. Higher-resource languages tend to be longer probably because the Reddit discussions are more in-depth. Nonetheless, the tokenizer is highly language-dependent and the length is not strictly comparable among languages.

\begin{figure}[t!]
\centering
\includegraphics[width=0.45\textwidth]{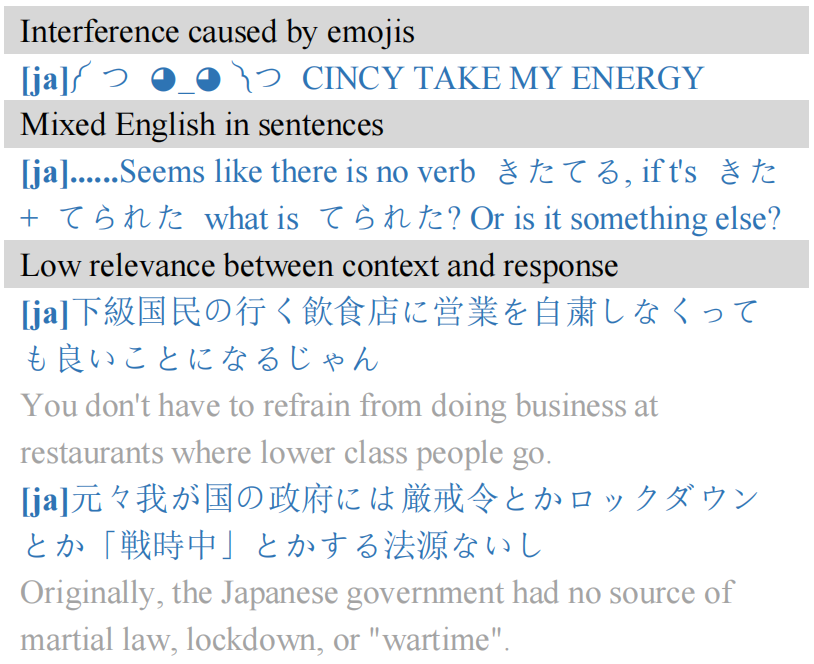}        
\caption{\small Corpus noise examples.}
\label{fig:noise}
 
\end{figure}

\noindent\textbf{Human Evaluations} We perform human evaluation on 10 of the languages in the dataset: English, Korean, French, Japanese, Italian, Spanish, Chinese, Tagalog, Hindi and Marathi. We randomly sample 100 conversations for each of them, then ask humans to evaluate (1) if the conversation is in the correct language (correct). (2) if the response is a natural sentence (natural), (3) if the response is relevant with the context (relevant) and (4) if the response is diverse (diverse). The correctness is judged as a binary score and the rest are in the 5-class Likert scale from 5 (best) to 1 (worst). The latter 3 are only scored when the conversation is in the right language. Each conversation is annotated by two native speakers and the averaged score is reported\footnote{We measured inter-annotator agreement using Cohen’s
kappa coefficient. Most scores were within the 0.6-0.9 range,
showing high agreement between the evaluators.} in Table~\ref{tab:human_ground}. We can see the quality is good for most languages. As some noise examples from the corpus showed in Figure~\ref{fig:noise}, we can see that languages can be misclassified when mixed with emojis or other languages. It can also be judged as low relevance when the context and response have no connection without further information.

\begin{table}[!t]
 	 \small
    	\centering	
     	\scalebox{0.95}
    		{
    	\begin{tabular}{c|c|c|c|c} 
    	\hline
    	\textbf{Language}&\textbf{Correct}&\textbf{Natural}&\textbf{Relevant}&\textbf{Diverse}\\
    	English & 0.95  & 4.69  & 4.18  & 4.62 \\
    French & 0.92  & 4.31  & 4.05  & 4.33 \\
    Japanese & 0.96  & 4.39  & 4.11  & 4.41 \\
    Italian & 0.95  & 4.27  & 3.96  & 4.31 \\
    Spanish & 0.92  & 4.32  & 3.89  & 4.34 \\
    Chinese & 0.92  & 4.34  & 4.32  & 4.39 \\
    Tagalog & 0.95  & 4.39  & 3.98  & 4.07 \\
    Korean & 0.98  & 4.59  & 4.63  & 4.54 \\
    Hindi & 0.88  & 4.17  & 4.25  & 4.04 \\
    Marathi & 0.85  & 4.20  & 3.82  & 4.43 \\

\hline
    	\end{tabular}

}
    	\caption{\label{tab:human_ground}\small Human evaluation of sampled conversations.} \vspace{-3mm}
\end{table}

\section{Models}
We experimented with two types of models. The first type is built on the pre-trained language model mT5~\cite{xue2021mt5}, which includes language-specific knowledge but has not been tailored for dialogue generation. The second type is built on a well developed English chatbot DialoGPT, which is trained intensively on conversational data but focuses only on English~\cite{zhang2020dialogpt}.

\noindent\textbf{mT5-based}
MT5 is a multilingual language model pre-trained on 1 trillion tokens covering 101 languages. All the 46 languages in \textsc{mDIA} are covered in the training corpus of mT5 so it has knowledge about all the languages encoded in its weights. We use its base version (600M parameters) in our experiments and train two different versions of models: (1) Finetune an mT5 model for every individual language (\emph{mono-mT5}) and (2) mix the training corpus from all languages and train a single model (\emph{multi-mT5}).
The latter has been a common choice for multilingual generation tasks~\cite{johnson2017google,hasan2021xl} but has not been tried in dialogue generation yet. By mixing all the training corpus, similar languages can take advantage of positive transfer from each other.

\begin{figure}[t!]
 
\centering
\includegraphics[width=0.45\textwidth]{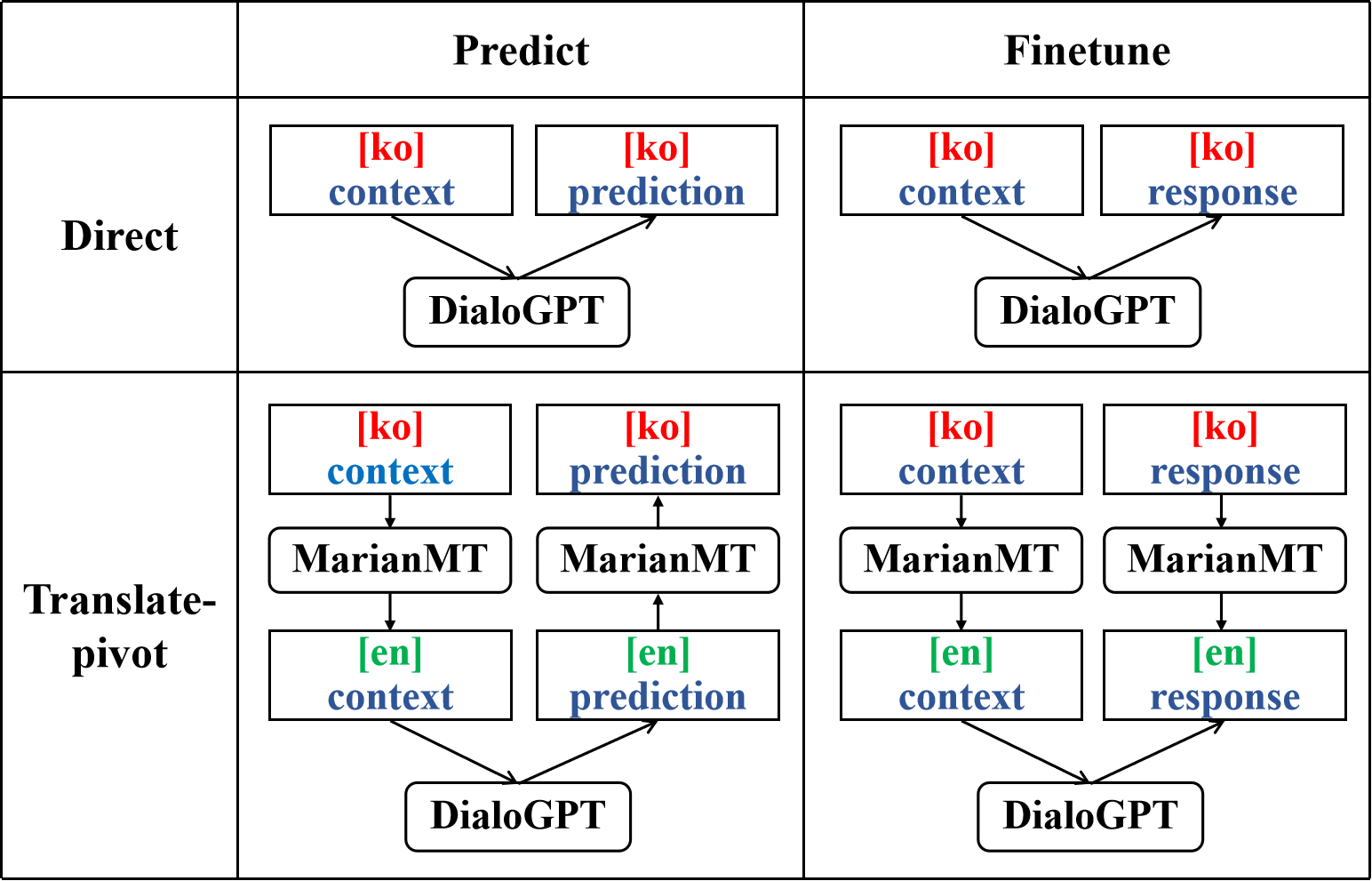}        
\caption{\small Illustration of 4 DialoGPT Variants.}
\label{fig:dialogpt_model}
 
\end{figure}

\noindent\textbf{DialoGPT-based}
DialoGPT~\cite{zhang2020dialogpt} is a pre-trained dialogue system for English. It is trained on 147M conversation-like exchanges collected from the Reddit posts of 2005 through 2017. We use its large version (762M, the closest scale with mt5-based) and train the following four versions of models: (1) \emph{Direct\_predict}. Directly applies DialoGPT to generate the response for a given context. (2) \emph{Direct\_finetune}. Directly finetune DialoGPT on the target language. Then we apply \emph{direct\_predict} on the finetuned DialoGPT. (3) \emph{Translate-pivot\_predict}. The context is translated into English, fed into DialoGPT to get the English response, then translated back to the target langauge and (4) \emph{Translate-pivot\_finetune}. Translate the training corpus into English and finetun DialoGPT on it. Then we apply \emph{Translate-pivot\_predict} on the finetuned DialoGPT.
Figure~\ref{fig:dialogpt_model} illustrate the above 4 variants.

\emph{Direct\_predict} and \emph{direct\_finetune} make straightforward reuse of the vocabulary and model weight from DialoGPT. Even though DialoGPT targets English, its training corpus is inevitably mixed with other languages and therefore has seen many of the foreign words.\footnote{It adopted filtering mechanisms to keep only English sentences to the largest extend, so foreign languages will be very limited in their training corpus.} However, the low frequency of foreign language in its corpus makes this direct adaptation difficult especially for languages with different scripts~\cite{adewumi2021sm}.

\emph{Translate-pivot\_predict} leverages a translator\footnote{We use MarianMT~\cite{junczys2018marian} as our translator through this paper.} as the bridge to break the vocabulary and grammar mismatch among languages. However, it strongly depends on the translation quality. Furthermore, even if the translator can work perfectly, it might still generate unnaturally since different languages have their own styles and habits in daily conversations~\cite{moorjani1988semiotic}.

\emph{Finetune\_translate-pivot} can adapt DialoGPT to the translation errors and conversational styles in the target language by finetuning it on the translated corpus of the target language. However, it relies on the cycle consistency~\cite{he2016dual} of the translator. If the translator is not cycle consistent, even if DialoGPT adapts perfectly to the translated English, the generated response might still be bad when translated back to the target language.

We term the first two models as zero-shot DialoGPT (\textit{0-DGPT}) and fine-tuned DialoGPT (\textit{ft-DGPT}), the last two models as zero-shot dialoGPT + MarianMT (\textit{0-DGPT+mt}) and fine-tuned DialoGPT + MarianMT (\textit{ft-DGPT+mt}).

\section{Experiments}
\subsection{Experiment Setup}
\noindent\textbf{Detailed Configurations} All models are implemented based on the Transformers-PyTorch library~\cite{wolf2020transformers}, and trained with maximum context and response with 64 BPE tokens, and longer examples are truncated. We apply the Adafactor optimizer~\cite{shazeer2018adafactor} and set the weight decay as 0.1. Models are trained with the batch size of 64 with a maximum of 30 epochs with EarlyStopping callbacks to restrain over fitting. The learning rate is chosen from $[1e-4, 5e-5, 3e-5, 1e-5]$ and the warm up step is chosen from $[20\%,50\%, 100\%, 200\%]$ of one training epoch. The hyperparameters with the lowest validation loss are used. When decoding, we apply nucleus sampling~\cite{holtzman2019curious} with $p=0.7$, no-repeat-ngram=2 to promote the diversity of generation.

\noindent\textbf{Evaluation Metrics}
We evaluate the quality of generated responses with sacreBLEU~\cite{papineni2002bleu} and BertScore~\cite{Zhang*2020BERTScore}. Both of them are realized on multilingual tokenizers and fixed parameters, ensuring the reproducibility and the comparability of our benchmark. We also use Entropy~\cite{zhang2018generating} and Dist-n~\cite{li2016diversity,su2018dialogue} to evaluate lexical diversity. We also apply human evaluation for sampled generated responses the similar way as we evaluate the dataset.

\noindent\textbf{High-resource vs. Low-resource} We group the whole 45 non-English languages into 17 relatively high-resource languages and 28 low-resource languages. By high-resource, we refer to languages with 12k conversations and have bidirectional translation systems with English available in MarianMT. The others are treated as low-resource languages. The high/low-resource here only reflect the situation in Reddit and MarianMT and do not necessarily correlate with the resource availability in other domains or the number of actual speakers.

\subsection{Results on High-Resource Languages}
\begin{figure*}[t!]
\centering
\includegraphics[width=0.8\textwidth]{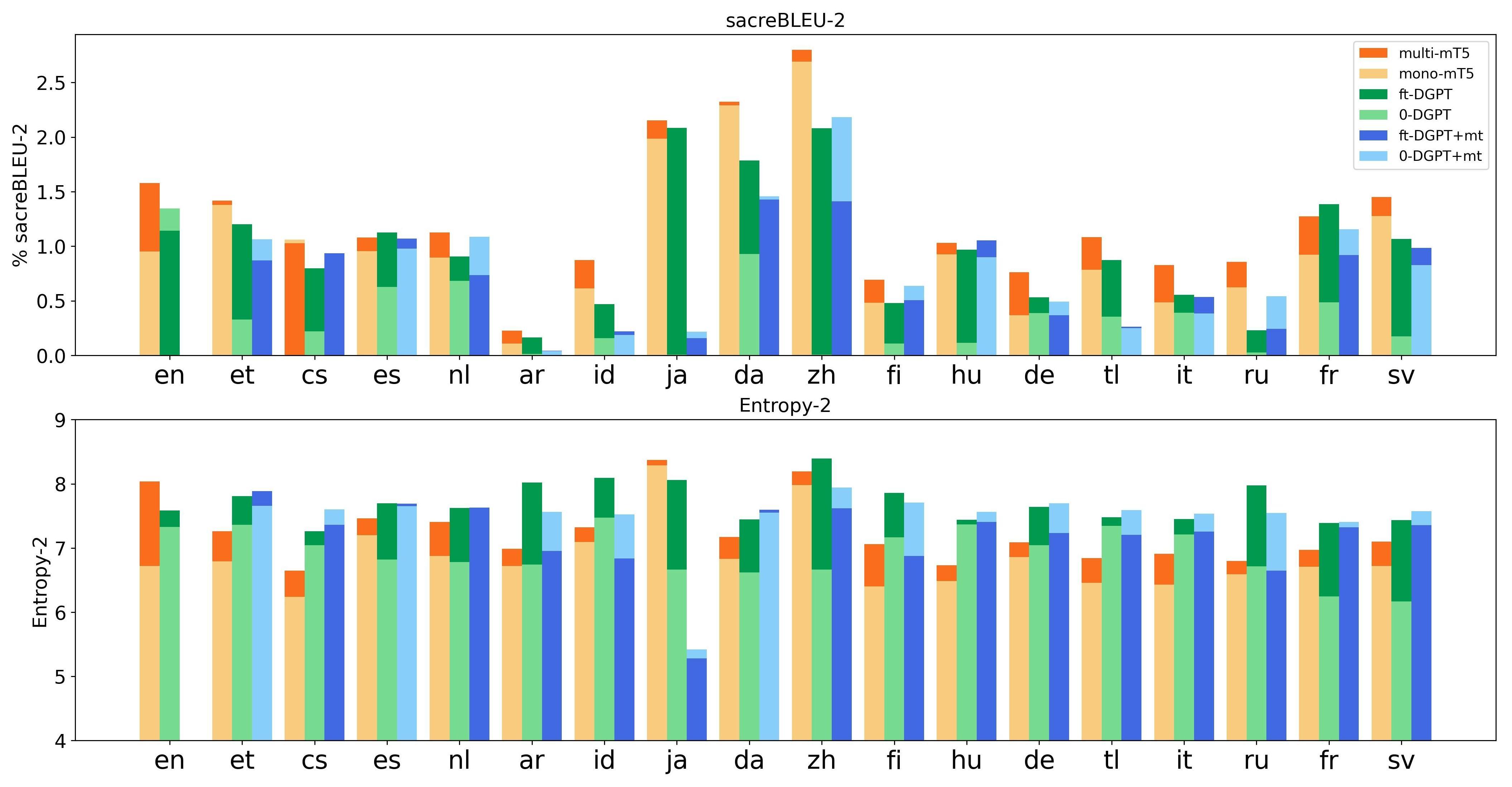}        
\caption{\small Performance of high-resource languages with different models}
\label{fig:metrics17}
 
\end{figure*}

We show the exact metric scores for best results among the 6 models in Table~\ref{tab:metrics17}. The comparison of the 6 models are visualized in Figure~\ref{fig:metrics17}.
\begin{table*}
\centering
\small
\resizebox{\textwidth}{37mm}{
\begin{tabular}{lcrrrrrrrrrrr}
\hline \multirow{2}{*}{\small\textbf{Language}} & \multirow{2}{*}{\small\textbf{Family}} & \multicolumn{4}{c}{\small\textbf{sacreBLEU}} & \multicolumn{3}{c}{\small\textbf{Bert Score}} & \multicolumn{2}{c}{\small\textbf{Distinct}} & \multicolumn{2}{c}{\small\textbf{Entropy}} \\
&&score&B-2&B-4&BP&Pre&Rec&F1&D-2&D-4&E-2&E-4\\\hline
\textbf{English} & Germanic & \cellcolor[rgb]{ 1,  .851,  .396}0.96\% & \cellcolor[rgb]{ 1,  .851,  .396}1.58\% & \cellcolor[rgb]{ 1,  .851,  .396}0.38\% & \cellcolor[rgb]{ .663,  .804,  .565}0.82 & \cellcolor[rgb]{ .773,  .871,  .71}85.7\% & \cellcolor[rgb]{ .663,  .804,  .565}83.9\% & \cellcolor[rgb]{ .773,  .871,  .71}84.6\% & \cellcolor[rgb]{ 1,  .851,  .396}0.40 & \cellcolor[rgb]{ .663,  .804,  .565}0.65 & \cellcolor[rgb]{ 1,  .851,  .396}8.04 & \cellcolor[rgb]{ 1,  .851,  .396}9.33 \\
Estonian & Uralic & \cellcolor[rgb]{ 1,  .851,  .396}0.65\% & \cellcolor[rgb]{ 1,  .851,  .396}1.42\% & \cellcolor[rgb]{ 1,  .949,  .796}0.17\% & \cellcolor[rgb]{ .557,  .667,  .859}0.87 & \cellcolor[rgb]{ .851,  .886,  .953}66.7\% & \cellcolor[rgb]{ 1,  .851,  .396}64.2\% & \cellcolor[rgb]{ 1,  .851,  .396}65.1\% & \cellcolor[rgb]{ 1,  .949,  .796}0.44 & \cellcolor[rgb]{ .663,  .804,  .565}0.69 & \cellcolor[rgb]{ .557,  .667,  .859}7.89 & \cellcolor[rgb]{ .663,  .804,  .565}9.35 \\
Czech & Slavic & \cellcolor[rgb]{ 1,  .851,  .396}0.54\% & \cellcolor[rgb]{ .773,  .871,  .71}1.06\% & \cellcolor[rgb]{ .557,  .667,  .859}0.06\% & \cellcolor[rgb]{ .663,  .804,  .565}0.90 & \cellcolor[rgb]{ .851,  .886,  .953}65.4\% & \cellcolor[rgb]{ .663,  .804,  .565}63.5\% & \cellcolor[rgb]{ .557,  .667,  .859}64.2\% & \cellcolor[rgb]{ .851,  .886,  .953}0.39 & \cellcolor[rgb]{ .851,  .886,  .953}0.65 & \cellcolor[rgb]{ .851,  .886,  .953}7.60 & \cellcolor[rgb]{ .663,  .804,  .565}9.07 \\
Spanish & Romance & \cellcolor[rgb]{ 1,  .851,  .396}0.69\% & \cellcolor[rgb]{ .663,  .804,  .565}1.13\% & \cellcolor[rgb]{ .851,  .886,  .953}0.08\% & \cellcolor[rgb]{ .557,  .667,  .859}0.96 & \cellcolor[rgb]{ .851,  .886,  .953}67.5\% & \cellcolor[rgb]{ 1,  .851,  .396}65.0\% & \cellcolor[rgb]{ .557,  .667,  .859}65.8\% & \cellcolor[rgb]{ .851,  .886,  .953}0.40 & \cellcolor[rgb]{ .663,  .804,  .565}0.68 & \cellcolor[rgb]{ .663,  .804,  .565}7.70 & \cellcolor[rgb]{ 1,  .851,  .396}9.58 \\
Dutch & Germanic & \cellcolor[rgb]{ 1,  .851,  .396}0.70\% & \cellcolor[rgb]{ 1,  .851,  .396}1.13\% & \cellcolor[rgb]{ .851,  .886,  .953}0.20\% & \cellcolor[rgb]{ .663,  .804,  .565}0.96 & \cellcolor[rgb]{ .773,  .871,  .71}67.6\% & \cellcolor[rgb]{ 1,  .851,  .396}65.0\% & \cellcolor[rgb]{ 1,  .851,  .396}65.8\% & \cellcolor[rgb]{ .851,  .886,  .953}0.38 & \cellcolor[rgb]{ .663,  .804,  .565}0.68 & \cellcolor[rgb]{ .851,  .886,  .953}7.63 & \cellcolor[rgb]{ .663,  .804,  .565}9.50 \\
Arabic & Arabic & \cellcolor[rgb]{ .663,  .804,  .565}0.18\% & \cellcolor[rgb]{ 1,  .851,  .396}0.23\% & \cellcolor[rgb]{ .663,  .804,  .565}0.03\% & \cellcolor[rgb]{ .663,  .804,  .565}1.00 & \cellcolor[rgb]{ .773,  .871,  .71}67.3\% & \cellcolor[rgb]{ .663,  .804,  .565}66.4\% & \cellcolor[rgb]{ .663,  .804,  .565}66.3\% & \cellcolor[rgb]{ .663,  .804,  .565}0.39 & \cellcolor[rgb]{ .663,  .804,  .565}0.72 & \cellcolor[rgb]{ .663,  .804,  .565}8.02 & \cellcolor[rgb]{ .663,  .804,  .565}9.31 \\
Indonesian & Malayo-Polyn & \cellcolor[rgb]{ 1,  .851,  .396}0.47\% & \cellcolor[rgb]{ 1,  .851,  .396}0.87\% & \cellcolor[rgb]{ 1,  .851,  .396}0.07\% & \cellcolor[rgb]{ .557,  .667,  .859}0.98 & \cellcolor[rgb]{ 1,  .851,  .396}66.6\% & \cellcolor[rgb]{ .663,  .804,  .565}64.8\% & \cellcolor[rgb]{ 1,  .851,  .396}65.5\% & \cellcolor[rgb]{ 1,  .949,  .796}0.44 & \cellcolor[rgb]{ .663,  .804,  .565}0.76 & \cellcolor[rgb]{ .663,  .804,  .565}8.09 & \cellcolor[rgb]{ .663,  .804,  .565}9.65 \\
\textbf{Japanese} & Japonic & \cellcolor[rgb]{ .663,  .804,  .565}2.14\% & \cellcolor[rgb]{ 1,  .851,  .396}2.15\% & \cellcolor[rgb]{ .663,  .804,  .565}0.89\% & \cellcolor[rgb]{ .557,  .667,  .859}1.00 & \cellcolor[rgb]{ 1,  .851,  .396}66.0\% & \cellcolor[rgb]{ 1,  .851,  .396}64.5\% & \cellcolor[rgb]{ 1,  .851,  .396}65.2\% & \cellcolor[rgb]{ .663,  .804,  .565}0.50 & \cellcolor[rgb]{ 1,  .851,  .396}0.66 & \cellcolor[rgb]{ 1,  .851,  .396}8.37 & \cellcolor[rgb]{ 1,  .851,  .396}9.09 \\
Danish & Germanic & \cellcolor[rgb]{ 1,  .851,  .396}1.62\% & \cellcolor[rgb]{ 1,  .851,  .396}2.32\% & \cellcolor[rgb]{ .773,  .871,  .71}0.38\% & \cellcolor[rgb]{ .663,  .804,  .565}1.00 & \cellcolor[rgb]{ .851,  .886,  .953}68.5\% & \cellcolor[rgb]{ 1,  .851,  .396}65.4\% & \cellcolor[rgb]{ 1,  .851,  .396}66.1\% & \cellcolor[rgb]{ .851,  .886,  .953}0.39 & \cellcolor[rgb]{ .851,  .886,  .953}0.67 & \cellcolor[rgb]{ .557,  .667,  .859}7.60 & \cellcolor[rgb]{ .663,  .804,  .565}9.40 \\
\textbf{Chinese} & Chinese & \cellcolor[rgb]{ 1,  .851,  .396}1.52\% & \cellcolor[rgb]{ 1,  .851,  .396}2.80\% & \cellcolor[rgb]{ .663,  .804,  .565}0.23\% & \cellcolor[rgb]{ .663,  .804,  .565}1.00 & \cellcolor[rgb]{ 1,  .949,  .796}58.6\% & \cellcolor[rgb]{ .663,  .804,  .565}57.1\% & \cellcolor[rgb]{ 1,  .949,  .796}57.7\% & \cellcolor[rgb]{ .851,  .886,  .953}0.53 & \cellcolor[rgb]{ .663,  .804,  .565}0.77 & \cellcolor[rgb]{ .663,  .804,  .565}8.39 & \cellcolor[rgb]{ 1,  .851,  .396}9.71 \\
Finnish & Uraric & \cellcolor[rgb]{ 1,  .851,  .396}0.30\% & \cellcolor[rgb]{ 1,  .851,  .396}0.69\% & \cellcolor[rgb]{ .663,  .804,  .565}0.04\% & \cellcolor[rgb]{ .557,  .667,  .859}1.00 & \cellcolor[rgb]{ .851,  .886,  .953}66.9\% & \cellcolor[rgb]{ .557,  .667,  .859}63.5\% & \cellcolor[rgb]{ .851,  .886,  .953}64.7\% & \cellcolor[rgb]{ .851,  .886,  .953}0.44 & \cellcolor[rgb]{ .663,  .804,  .565}0.71 & \cellcolor[rgb]{ .663,  .804,  .565}7.86 & \cellcolor[rgb]{ .663,  .804,  .565}9.41 \\
Hungarian & Uralic & \cellcolor[rgb]{ 1,  .851,  .396}0.59\% & \cellcolor[rgb]{ .557,  .667,  .859}1.05\% & \cellcolor[rgb]{ 1,  .851,  .396}0.05\% & \cellcolor[rgb]{ 1,  .851,  .396}1.00 & \cellcolor[rgb]{ .851,  .886,  .953}66.2\% & \cellcolor[rgb]{ .663,  .804,  .565}63.5\% & \cellcolor[rgb]{ .851,  .886,  .953}64.4\% & \cellcolor[rgb]{ 1,  .949,  .796}0.43 & \cellcolor[rgb]{ .851,  .886,  .953}0.65 & \cellcolor[rgb]{ .851,  .886,  .953}7.56 & \cellcolor[rgb]{ .663,  .804,  .565}9.29 \\
German & Germanic & \cellcolor[rgb]{ 1,  .851,  .396}0.53\% & \cellcolor[rgb]{ 1,  .851,  .396}0.76\% & \cellcolor[rgb]{ 1,  .851,  .396}0.08\% & \cellcolor[rgb]{ .557,  .667,  .859}1.00 & \cellcolor[rgb]{ 1,  .851,  .396}65.8\% & \cellcolor[rgb]{ 1,  .851,  .396}64.3\% & \cellcolor[rgb]{ 1,  .851,  .396}65.0\% & \cellcolor[rgb]{ .851,  .886,  .953}0.41 & \cellcolor[rgb]{ .851,  .886,  .953}0.67 & \cellcolor[rgb]{ .851,  .886,  .953}7.70 & \cellcolor[rgb]{ .663,  .804,  .565}9.17 \\
Tagalog & Malayo-Polyn & \cellcolor[rgb]{ 1,  .851,  .396}0.60\% & \cellcolor[rgb]{ 1,  .851,  .396}1.09\% & \cellcolor[rgb]{ .663,  .804,  .565}0.10\% & \cellcolor[rgb]{ .557,  .667,  .859}0.81 & \cellcolor[rgb]{ .663,  .804,  .565}67.1\% & \cellcolor[rgb]{ .663,  .804,  .565}65.2\% & \cellcolor[rgb]{ .663,  .804,  .565}66.1\% & \cellcolor[rgb]{ 1,  .949,  .796}0.41 & \cellcolor[rgb]{ .663,  .804,  .565}0.66 & \cellcolor[rgb]{ .851,  .886,  .953}7.59 & \cellcolor[rgb]{ .663,  .804,  .565}9.24 \\
Italian & Romance & \cellcolor[rgb]{ 1,  .851,  .396}0.59\% & \cellcolor[rgb]{ 1,  .851,  .396}0.83\% & \cellcolor[rgb]{ 1,  .851,  .396}0.11\% & \cellcolor[rgb]{ .663,  .804,  .565}0.99 & \cellcolor[rgb]{ .851,  .886,  .953}65.6\% & \cellcolor[rgb]{ .663,  .804,  .565}63.9\% & \cellcolor[rgb]{ .851,  .886,  .953}64.4\% & \cellcolor[rgb]{ 1,  .949,  .796}0.40 & \cellcolor[rgb]{ .851,  .886,  .953}0.65 & \cellcolor[rgb]{ .851,  .886,  .953}7.53 & \cellcolor[rgb]{ .663,  .804,  .565}9.06 \\
Russian & Slavic & \cellcolor[rgb]{ 1,  .949,  .796}0.22\% & \cellcolor[rgb]{ 1,  .851,  .396}0.86\% & \cellcolor[rgb]{ .773,  .871,  .71}0.01\% & \cellcolor[rgb]{ .557,  .667,  .859}1.00 & \cellcolor[rgb]{ .851,  .886,  .953}66.0\% & \cellcolor[rgb]{ .663,  .804,  .565}64.1\% & \cellcolor[rgb]{ 1,  .851,  .396}64.6\% & \cellcolor[rgb]{ .851,  .886,  .953}0.40 & \cellcolor[rgb]{ .663,  .804,  .565}0.75 & \cellcolor[rgb]{ .663,  .804,  .565}7.98 & \cellcolor[rgb]{ .663,  .804,  .565}9.40 \\
French & Romance & \cellcolor[rgb]{ .663,  .804,  .565}1.01\% & \cellcolor[rgb]{ .663,  .804,  .565}1.39\% & \cellcolor[rgb]{ .663,  .804,  .565}0.24\% & \cellcolor[rgb]{ .557,  .667,  .859}0.88 & \cellcolor[rgb]{ .851,  .886,  .953}67.3\% & \cellcolor[rgb]{ 1,  .851,  .396}65.1\% & \cellcolor[rgb]{ 1,  .851,  .396}66.0\% & \cellcolor[rgb]{ .851,  .886,  .953}0.33 & \cellcolor[rgb]{ .851,  .886,  .953}0.61 & \cellcolor[rgb]{ .851,  .886,  .953}7.41 & \cellcolor[rgb]{ .663,  .804,  .565}9.14 \\
Swedish & Germanic & \cellcolor[rgb]{ 1,  .851,  .396}0.77\% & \cellcolor[rgb]{ 1,  .851,  .396}1.45\% & \cellcolor[rgb]{ 1,  .851,  .396}0.13\% & \cellcolor[rgb]{ .663,  .804,  .565}0.90 & \cellcolor[rgb]{ .851,  .886,  .953}67.9\% & \cellcolor[rgb]{ 1,  .851,  .396}64.9\% & \cellcolor[rgb]{ .557,  .667,  .859}66.0\% & \cellcolor[rgb]{ .851,  .886,  .953}0.38 & \cellcolor[rgb]{ .851,  .886,  .953}0.67 & \cellcolor[rgb]{ .851,  .886,  .953}7.57 & \cellcolor[rgb]{ .663,  .804,  .565}9.26 \\
\hline
\end{tabular}
}
\caption{\label{tab:metrics17} \small Best results from
\colorbox[rgb]{ 1,  .949,  .8}{Mono-mT5},
\colorbox[rgb]{ 1,  .851,  .4}{Multi-mT5},
\colorbox[rgb]{ .886,  .937,  .855}{0-DGPT}, 
\colorbox[rgb]{ .663,  .816,  .557}{ft-DGPT}, 
\colorbox[rgb]{ .839,  .863, 
.894}{0-DGPT+mt} and 
\colorbox[rgb]{ .557,  .663,  .859}{ft-DGPT+mt} for high-resource languages. sacreBLEU, Bert Score, Distinct-2/4 and Entropy-2/4 are reported. It should be noted that English, Chinese, and Japanese are bloded because of special settings mentioned in Section 5.2, showing that score comparisons across languages are meaningless. The detailed results of more than 400 sets of experimental data are in the appendix.}
\end{table*}

\noindent\textbf{Best Results}
In Table~\ref{tab:metrics17}, only the result from the best-performed model for every metric is reported.
We can see that BertScore on English are significantly higher than on other languages, this comes from two reasons: (1)~DialoGPT has a big gap for generation quality on English, and (2)~BertScore use a specific model different from other languages. The lower BertScore in Chinese is for the same reason. Besides, similar reason on Chinese and Japanese's outstanding sacreBLEU scores is that they use specific tokenizers. The analysis above and difference in tokenizers of languages\cite{rehbein2007treebank} suggest that score comparisons across languages are \textbf{meaningless}.
For most languages, \emph{mT-5}-based models perform better on sacreBLEU and BertScore but \emph{DialoGPT}-based models perform better on diversity.

\noindent\textbf{mT5 vs. DialoGPT}
We can see from Figure~\ref{fig:metrics17} that mT5-based models consistently perform better on sacreBLEU and mostly on BertScore too. For diversity-based metrics, however, DialoGPT-based models are usually better. Considering that mT5-based models have the best brevity-penalty for all languages except for Japanese. This suggests \emph{DialoGPT generates shorter responses but with higher lexical diversity}. This could be ascribed to their dialogue-specific pre-training during which has seen many more conversation examples. MT5 has only been finetuned on a limited amount of conversational data, so it has more chances of overfitting to generic patterns. Its high sacreBLEU score also suggests it tends to find safe solutions for higher lexical overlap. 

\noindent\textbf{Mono vs. Multi}
\emph{Multi-mT5 outperforms mono-mT5 in most cases}, both for similarity-based metrics sacreBLEU/BertScore and diversity-based metrics distinct-i/entropy-i. However, their difference is usually small. Only in a few languages like English and Italian its advantage becomes visible. We hypothesize it will be more useful when more similar languages are mixed together and larger amounts of training data is available. Another possible reason is that dialogues are more language-specific. The same chatting patterns might not be easily transferable to another language even for closely related languages. This is in contrast with other tasks like summarization, where the goal of the task is clearly defined and easier to transfer among languages~\cite{hasan2021xl}.

\noindent\textbf{Use of Translate-pivot}
\emph{Translate-pivot helps for most languages especially in the zero-shot setting}, but its effects are highly language dependent. Intuitively it is helpful if the language exists very rarely in the training corpus of DialoGPT and its translation quality with English is high. For languages like Japanese, Estonian, Hungarian and Tagalog, the zero-shot DialoGPT performs very poorly. After finetuning, however, the performance is greatly boosted and outperforms translate-pivot methods. This should be because they occur frequently in Reddit themselves (See Figure~\ref{fig:distribution}). The vocabulary set of DialoGPT has included many of their words so that it can be quickly adapted to the target language under 10k training examples. Besides, the translation qualities for them are also unsatisfactory due to the paucity of parallel resources and language differences~\cite{bugliarello2020s}. The poor translation quality and the error propagation in the pivot process might outweigh its advantage of bridging the vocabulary and grammar gap.

\noindent\textbf{Direct vs. Finetuned}
Without exception, finetuning on the training data of the target language always helps \emph{when the translate-pivot is not applied}. After applying the translate-pivot, however, finetuning on the training data can sometimes negatively affect the performance (e.g., for Chinese and Dutch). The reason could be due to the poor translation quality plus the lack of cycle-consistency in the translation system (the same sentence, after being translated to English then translated back, is far from the original sentence). When the cycle-consistency is poor, finetuning could mislead the model.

\noindent\textbf{Human Evaluations} We conduct human evaluation for the best-performing system (based on the win-most system on sacreBLEU and BERTScore in Table~\ref{tab:metrics17}) on 9 languages. For each language, we randomly sample 100 context-response pairs for evaluation. 
We use Sensibleness and Specificity Average (SSA)\cite{adiwardana2020towards}, which captures key elements of conversations, including whether the bot's respond is make sense in context and whether the response is specific given the context.
Results are shown in Table~\ref{tab:human_high}. We see that Scores \textbf{across} languages do not correlate well with automatic metrics. For example, Japanese have the highest sacreBLEU score, but it has decent Sensibleness scores to ohers in Table~\ref{tab:metrics17}. This further suggests that sacreBLEU and BERTScore might be helpful for in-language comparison, but are \textbf{not comparable across languages}.

\begin{table}[!t]
 	 \small
    	\centering	
    		{
    	\begin{tabular}{c|c|c|c} 
    	\hline
    	\textbf{Language}&\textbf{Sensibleness}&\textbf{Specificity}&\textbf{SSA}\\
\rowcolor[rgb]{ 1,  .851,  .396} English & 0.72  & 0.6   & 0.66 \\
\rowcolor[rgb]{ 1,  .851,  .396} Chinese & 0.63  & 0.66  & 0.65 \\
\rowcolor[rgb]{ 1,  .851,  .396} Spanish & 0.62  & 0.57  & 0.60 \\
\rowcolor[rgb]{ .663,  .804,  .565} French & 0.67  & 0.6   & 0.64 \\
\rowcolor[rgb]{ 1,  .851,  .396} German & 0.64  & 0.55  & 0.60 \\
\rowcolor[rgb]{ 1,  .851,  .396} Japanese & 0.62  & 0.57  & 0.60 \\
\rowcolor[rgb]{ 1,  .851,  .396} Italian & 0.62  & 0.58  & 0.60 \\
\rowcolor[rgb]{ .663,  .804,  .565} Tagalog & 0.59  & 0.63  & 0.61 \\
\rowcolor[rgb]{ 1,  .851,  .396} Dutch & 0.63  & 0.54  & 0.59 \\

\hline
    	\end{tabular}

}
    	\caption{\label{tab:human_high}\small Human evaluation for high-resource languages.} \vspace{-3mm}
\end{table}
\begin{figure*}[t!]
\centering
\includegraphics[width=0.8\textwidth]{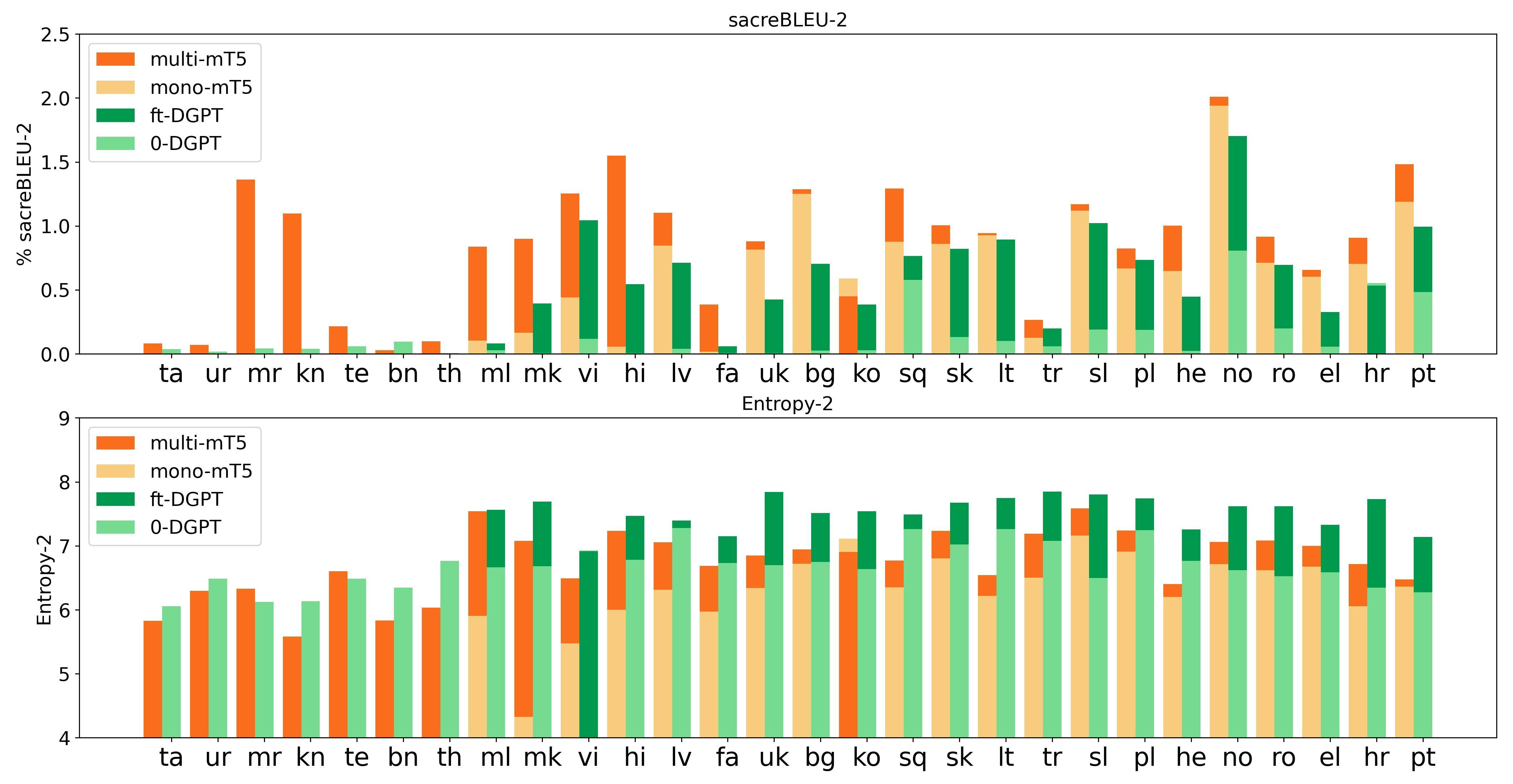}        
\caption{\small Performance of low-resource languages with different models.}
\label{fig:metrics35}
 
\end{figure*}
\subsection{Results on Low-Resource Languages}
For low-resource languages, as no bidirectional translator exists for many of them, we can only apply 4 out of the 6 proposed models on them (excluding translate-pivot models). For the 9 languages with fewer than 1k conversations in total, we can only apply 2 models on them: zero-shot DialoGPT and multilingual mT5. The other 2 models must rely on language-specific training data which do not exist for them. We show the comparison on low-resource languages in Figure~\ref{fig:metrics35}.

\noindent\textbf{mT5 vs. DialoGPT}
The difference is similar as for high-resource languages, but the results are more consistent. \emph{mT5-based models win across all low-resource languages in sacreBLEU and BertScore but loses in diversity} (except for Thai in entropy-2). This suggests retraining a dialogue generator based on the general multilingual language model is preferred for low-resource languages. The English-centric dialogue generator might have only limited linguistic overlap with these low-resource languages so that its knowledge in dialogue generation is hard to be transferred.

\noindent\textbf{Mono vs. Multi}
\emph{Multi-mT5 has clear advantages for languages with very few training samples} like Malayalam, Macedonian and Vietnamese. For other languages with more training samples, their difference becomes small again. \emph{Multi-mT5 also usually has higher diversity than mono-mT5}. The difference is larger than in high-resource languages, suggesting the diversity of lexical expressions can be transferable among languages. With only very few training samples, mono-mT5 models could be overfit to generic expressions.

\begin{table}[!t]
 	 \small
    	\centering	
    		{
    	\begin{tabular}{c|c|c|c} 
    	\hline
    	\textbf{Language}&\textbf{Sensibleness}&\textbf{Specificity}&\textbf{SSA}\\
\rowcolor[rgb]{ 1,  .851,  .396} Portuguese & 0.49  & 0.45  & 0.47  \\
\rowcolor[rgb]{ 1,  .851,  .396} Korean & 0.38  & 0.42  & 0.40  \\
\rowcolor[rgb]{ 1,  .851,  .396} Ukrainian & 0.44  & 0.47  & 0.46  \\
\rowcolor[rgb]{ 1,  .851,  .396} Persian & 0.28  & 0.31  & 0.30  \\
\rowcolor[rgb]{ 1,  .851,  .396} Hindi & 0.45  & 0.48  & 0.47  \\
\rowcolor[rgb]{ 1,  .851,  .396} Urdu  & 0.24  & 0.32  & 0.28  \\
\rowcolor[rgb]{ 1,  .851,  .396} Marathi & 0.28  & 0.34  & 0.31  \\

\hline
    	\end{tabular}

}
    	\caption{\label{tab:human_low}\small Human evaluation for low-resource languages.} \vspace{-3mm}
\end{table}

\noindent\textbf{Human Evaluations}
We perform human evaluation for responses from the multi-mT5, the best overall model from the automatic evaluations. The results are shown in Table~\ref{tab:human_low}. The quality is clearly worse than those for high-resource languages. Urdu and Marathi have very poor scores for correctness because they are not in the training set of multi-mT5. The model tends to generate in similar languages that exist in the training corpus.

\noindent\textbf{Generation Examples}

\begin{figure}[t!]
\centering
\includegraphics[width=0.4\textwidth]{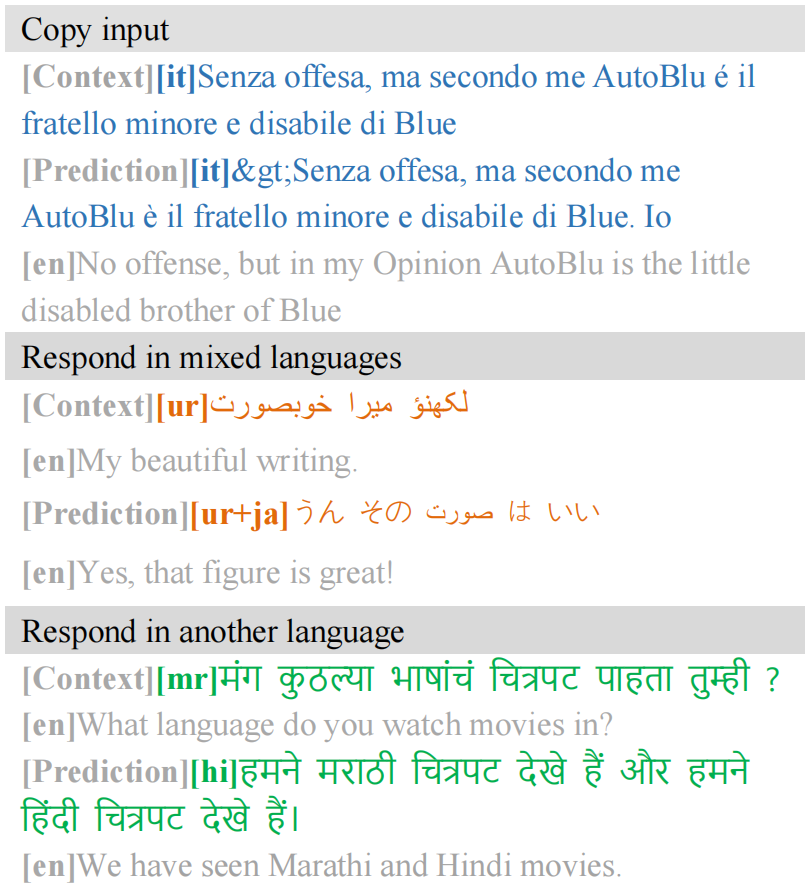}        
\caption{\small Generation Examples from Multi-MT5.}
\label{fig:gen_example}
 
\end{figure}

We show some generated examples from multi-mT5 in Figure~\ref{fig:gen_example}. We see the model sometimes simply copies the input as the response. It might not be able to fully learn the replying skills from the limited corpus. Interestingly, when generating responses for unseen languages, it is able to \emph{understand the context, but reply in other languages existing in the training corpus (sometimes even reply in mixed languages)}. Considering the huge differences in language families like ur and ja and their slim chance to mixed appear in the pre-training corpus, this is a promising result showing semantic alignment across languages in the encoder/decoder.

\section{Conclusion}
We present \textsc{mDIA}, the first benchmark for multilingual dialogue generation in 46 languages across 19 language families. We present baseline results based on both multilingual non-dialogue-focused, as well as English-centric, dialogue-focused pre-trained language model. Results show that the former performs better on sacreBLEU and BertScore but worse on diversity. It also show promising results on unseen languages. We hope the release of \textsc{mDIA} can call for more attention in this area and promote multilinguality in dialogue generation.

\section{Ethics Impact Analysis}
\label{sec:ethics}
Dialogue Generation Models are trained to mimic human conversations, so they will learn toxic knowledge from biased training data. Given the importance of the quality of our  multilingual dialogue generation corpus for relevant model generation, it is important to ensure our corpus is as innocuous as possible.

Many datasets focus on hate speech data\cite{huang2020multilingual} and gender bias\cite{JieyuZhao2020GenderBI} has been introduced to promote researches on suppressing toxic speech. \cite{EmilyDinan2019BuildIB} build a robust classifier of toxic language by introducing adversarial toxic data which fools existing classifiers. 

Although these works described above have greatly contributed to ethical impact research, the lack of low-resource data in these datasets is a huge problem in applying them to our work. Since we collected 100 toxic phrase in English and then translate them into other languages with Google translator to construct a toxic phrase vocabulary. We use a crude filter rule to ensure as little toxic speech in our corpus as possible that we remove comments if it contain any phrase in our toxic phrase vocabulary, removing 22.5\% dialogues.
We perform a human evaluation of the content of toxic speech in several languages in our corpus. The evaluation shows that our ethics impact filtering rule can clean toxic comments in our corpus effectively.

\bibliography{anthology,custom}
\bibliographystyle{acl_natbib}

\appendix


\begin{table*}[htbp]
\centering
\small
\caption{Statistics of different languages in \textsc{mDIA}}
\resizebox{.95\textwidth}{!}{
\begin{tabular}{ccccr}
\hline
\textbf{ISO} & \textbf{Language} & \textbf{Family} & \textbf{Script} & \textbf{\#Conversations} \\
\hline
    ar    & Arabic & Arabic & Arabic & 12000 \\
bg    & Bulgarian & Slavic & Cyrillic & 9429 \\
bn    & Bengali & Indo-Aryan & Eastern-Nagari & 298 \\
cs    & Czech & Slavic & Latin & 12000 \\
da    & Danish & Germanic & Latin & 12000 \\
de    & German & Germanic & Latin & 12000 \\
el    & Greek & Hellenic & Greek & 12000 \\
en    & English & Germanic & Latin & 12000 \\
es    & Spanish & Romance & Latin & 12000 \\
et    & Estonian & Uralic & Latin & 12000 \\
fa    & Persian & Iranian & Arabic & 3051 \\
fi    & Finnish & Uralic & Latin & 12000 \\
fr    & French & Romance & Latin & 12000 \\
he    & Hebrew & Semitic & Hebrew & 12000 \\
hi    & Hindi & Indo-Aryan & Devanagari & 2325 \\
hr    & Croatian & Slavic & Latin & 12000 \\
hu    & Hungarian & Uralic & Latin & 12000 \\
id    & Indonesian & Malayo-Polyn. & Latin & 12000 \\
it    & Italian & Romance & Latin & 12000 \\
ja    & Japanese & Japonic & Kanji; Kana & 12000 \\
kn    & Kannada & Tamil & Kannada & 167 \\
ko    & Korean & Koreanic & Hangul & 7060 \\
lt    & Lithuanian & Uralic & Latin & 9435 \\
lv    & Latvian & Uralic & Latin & 3697 \\
mk    & Macedonian & Slavic & Cyrillic & 2409 \\
ml    & Malayalam & Dravidian & Malayalam & 1610 \\
mr    & Marathi & Indo-Aryan & Devanagari & 165 \\
nl    & Dutch & Germanic & Latin & 12000 \\
no    & Norwegian & Germanic & Latin & 12000 \\
pl    & Polish & Slavic & Latin & 12000 \\
pt    & Portuguese & Romance & Latin & 12000 \\
ro    & Romanian & Romance & Latin & 12000 \\
ru    & Russian & Slavic & Cyrillic & 12000 \\
sk    & Slovak & Slavic & Latin & 9525 \\
sl    & Slovenian & Slavic & Latin & 12000 \\
sq    & Albanian & Albanian & Latin & 8797 \\
sv    & Swedish & Germanic & Latin & 12000 \\
ta    & Tamil & Dravidian & Tamil & 200 \\
te    & Telugu & Dravidian & Tamil & 486 \\
th    & Thai  & Kra-Dai & Thai  & 979 \\
tl    & Tagalog & Malayo-Polyn. & Latin & 12000 \\
tr    & Turkish & Turkic & Latin & 12000 \\
uk    & Ukrainian & Slavic & Cyrillic & 5548 \\
ur    & Urdu  & Indo-Aryan & Arabic & 402 \\
vi    & Vietnamese & Vietic & Latin & 3331 \\
zh    & Chinese & Chinese & Chinese & 12000 \\
\hline
\end{tabular}
}
\label{tab:lang_statistics}%
\end{table*}

\begin{table*}[htbp]
  \centering
  \small
  \caption{Performance of Multilingual mT5}
  \scalebox{1.1}
  {
    \begin{tabular}{lrrrrrrrrrrr}
\hline \multirow{2}{*}{\small\textbf{Language}} & \multicolumn{4}{c}{\small\textbf{sacreBLEU}} & \multicolumn{3}{c}{\small\textbf{BERTScore}} & \multicolumn{2}{c}{\small\textbf{Distinct}} & \multicolumn{2}{c}{\small\textbf{Entropy}} \\
&score&B-2&B-4&BP&Pre&Rec&F1&D-2&D-4&E-2&E-4\\
    \hline
mr    & 0.79\% & 1.36\% & 0.23\% & 0.79  & 67\%  & 65\%  & 66\%  & 0.39  & 0.64  & 6.33  & 7.29 \\
kn    & 0.67\% & 1.10\% & 0.13\% & 0.78  & 65\%  & 64\%  & 64\%  & 0.28  & 0.49  & 5.58  & 6.54 \\
ta    & 0.10\% & 0.08\% & 0.03\% & 0.69  & 63\%  & 60\%  & 61\%  & 0.26  & 0.46  & 5.83  & 6.92 \\
bn    & 0.04\% & 0.03\% & 0.01\% & 1.00  & 64\%  & 62\%  & 63\%  & 0.21  & 0.42  & 5.83  & 7.07 \\
ur    & 0.05\% & 0.07\% & 0.01\% & 0.74  & 64\%  & 63\%  & 63\%  & 0.24  & 0.47  & 6.29  & 7.48 \\
te    & 0.08\% & 0.21\% & 0.01\% & 0.57  & 64\%  & 62\%  & 63\%  & 0.28  & 0.58  & 6.60  & 8.17 \\
th    & 0.08\% & 0.10\% & 0.02\% & 1.00  & 59\%  & 57\%  & 58\%  & 0.15  & 0.27  & 6.03  & 6.92 \\
ml    & 0.34\% & 0.84\% & 0.06\% & 0.67  & 66\%  & 63\%  & 65\%  & 0.34  & 0.64  & 7.54  & 9.10 \\
hi    & 1.07\% & 1.55\% & 0.46\% & 0.71  & 68\%  & 66\%  & 67\%  & 0.28  & 0.60  & 7.23  & 9.08 \\
mk    & 0.68\% & 0.90\% & 0.09\% & 0.97  & 65\%  & 63\%  & 64\%  & 0.25  & 0.57  & 7.08  & 9.19 \\
fa    & 0.15\% & 0.39\% & 0.01\% & 0.84  & 67\%  & 67\%  & 67\%  & 0.20  & 0.53  & 6.69  & 8.69 \\
vi    & 0.74\% & 1.25\% & 0.23\% & 0.74  & 68\%  & 66\%  & 67\%  & 0.14  & 0.44  & 6.49  & 8.51 \\
lv    & 0.33\% & 1.10\% & 0.02\% & 0.84  & 65\%  & 63\%  & 64\%  & 0.22  & 0.56  & 7.05  & 9.08 \\
uk    & 0.31\% & 0.88\% & 0.02\% & 0.84  & 66\%  & 63\%  & 64\%  & 0.20  & 0.51  & 6.85  & 9.02 \\
ko    & 0.17\% & 0.45\% & 0.01\% & 0.75  & 69\%  & 66\%  & 67\%  & 0.21  & 0.51  & 6.90  & 8.42 \\
sq    & 1.23\% & 1.29\% & 0.59\% & 0.72  & 65\%  & 64\%  & 64\%  & 0.20  & 0.51  & 6.77  & 9.00 \\
bg    & 0.59\% & 1.29\% & 0.06\% & 0.84  & 66\%  & 64\%  & 65\%  & 0.22  & 0.53  & 6.94  & 9.03 \\
lt    & 0.23\% & 0.94\% & 0.01\% & 0.78  & 65\%  & 63\%  & 64\%  & 0.19  & 0.47  & 6.54  & 8.65 \\
sk    & 0.37\% & 1.01\% & 0.03\% & 0.78  & 65\%  & 63\%  & 63\%  & 0.24  & 0.60  & 7.23  & 9.23 \\
cs    & 0.54\% & 1.03\% & 0.06\% & 0.88  & 65\%  & 63\%  & 64\%  & 0.18  & 0.48  & 6.65  & 8.73 \\
et    & 0.65\% & 1.42\% & 0.13\% & 0.74  & 66\%  & 64\%  & 65\%  & 0.26  & 0.59  & 7.26  & 9.15 \\
ar    & 0.08\% & 0.23\% & 0.01\% & 0.75  & 67\%  & 65\%  & 66\%  & 0.21  & 0.55  & 6.99  & 8.92 \\
sl    & 0.41\% & 1.17\% & 0.03\% & 0.77  & 65\%  & 63\%  & 64\%  & 0.30  & 0.67  & 7.59  & 9.49 \\
he    & 0.56\% & 1.00\% & 0.07\% & 1.00  & 68\%  & 66\%  & 67\%  & 0.17  & 0.40  & 6.40  & 8.11 \\
ja    & 1.95\% & 2.15\% & 0.72\% & 0.94  & 66\%  & 64\%  & 65\%  & 0.48  & 0.66  & 8.37  & 9.09 \\
en    & 0.96\% & 1.58\% & 0.38\% & 0.66  & 85\%  & 84\%  & 84\%  & 0.40  & 0.64  & 8.04  & 9.33 \\
no    & 1.10\% & 2.01\% & 0.16\% & 0.85  & 68\%  & 66\%  & 66\%  & 0.22  & 0.52  & 7.06  & 9.16 \\
el    & 0.39\% & 0.66\% & 0.05\% & 0.81  & 66\%  & 64\%  & 65\%  & 0.21  & 0.50  & 7.00  & 8.90 \\
zh    & 1.52\% & 2.80\% & 0.19\% & 0.98  & 58\%  & 57\%  & 58\%  & 0.37  & 0.73  & 8.19  & 9.71 \\
da    & 1.62\% & 2.32\% & 0.37\% & 0.91  & 67\%  & 65\%  & 66\%  & 0.22  & 0.55  & 7.17  & 9.24 \\
tl    & 0.60\% & 1.08\% & 0.09\% & 0.81  & 66\%  & 65\%  & 65\%  & 0.22  & 0.53  & 6.85  & 8.88 \\
hu    & 0.59\% & 1.03\% & 0.05\% & 1.00  & 65\%  & 63\%  & 64\%  & 0.22  & 0.50  & 6.73  & 8.92 \\
id    & 0.47\% & 0.87\% & 0.07\% & 0.75  & 67\%  & 64\%  & 65\%  & 0.26  & 0.63  & 7.32  & 9.35 \\
fi    & 0.30\% & 0.69\% & 0.04\% & 0.69  & 66\%  & 63\%  & 65\%  & 0.24  & 0.57  & 7.06  & 9.06 \\
pl    & 0.51\% & 0.82\% & 0.09\% & 0.78  & 65\%  & 63\%  & 64\%  & 0.25  & 0.59  & 7.24  & 9.21 \\
ro    & 0.50\% & 0.92\% & 0.05\% & 0.82  & 64\%  & 63\%  & 63\%  & 0.24  & 0.55  & 7.08  & 9.16 \\
sv    & 0.77\% & 1.45\% & 0.12\% & 0.79  & 67\%  & 65\%  & 66\%  & 0.23  & 0.56  & 7.10  & 9.11 \\
hr    & 0.38\% & 0.91\% & 0.03\% & 0.78  & 65\%  & 63\%  & 64\%  & 0.21  & 0.50  & 6.71  & 8.88 \\
nl    & 0.70\% & 1.13\% & 0.12\% & 0.82  & 67\%  & 65\%  & 66\%  & 0.26  & 0.64  & 7.41  & 9.42 \\
it    & 0.59\% & 0.83\% & 0.11\% & 0.91  & 65\%  & 64\%  & 64\%  & 0.23  & 0.52  & 6.91  & 8.83 \\
de    & 0.53\% & 0.76\% & 0.08\% & 0.90  & 66\%  & 64\%  & 65\%  & 0.29  & 0.59  & 7.09  & 8.93 \\
ru    & 0.20\% & 0.86\% & 0.00\% & 1.00  & 65\%  & 64\%  & 65\%  & 0.21  & 0.54  & 6.80  & 9.03 \\
pt    & 0.95\% & 1.48\% & 0.15\% & 0.93  & 67\%  & 66\%  & 66\%  & 0.22  & 0.48  & 6.47  & 8.58 \\
fr    & 0.72\% & 1.27\% & 0.13\% & 0.77  & 67\%  & 65\%  & 66\%  & 0.21  & 0.50  & 6.97  & 8.86 \\
tr    & 0.16\% & 0.27\% & 0.04\% & 0.67  & 43\%  & 40\%  & 41\%  & 0.30  & 0.58  & 7.19  & 8.75 \\
es    & 0.69\% & 1.08\% & 0.08\% & 0.96  & 66\%  & 65\%  & 66\%  & 0.26  & 0.65  & 7.46  & 9.58 \\

    \hline
    \end{tabular}%
    }
  \label{tab:lang_multit5}%
\end{table*}%

\begin{table*}[htbp]
  \centering
  \small
  \caption{Performance of Monolingual mT5}
  \scalebox{1.1}
  {
    \begin{tabular}{lrrrrrrrrrrr}
\hline \multirow{2}{*}{\small\textbf{Language}} & \multicolumn{4}{c}{\small\textbf{sacreBLEU}} & \multicolumn{3}{c}{\small\textbf{BERTScore}} & \multicolumn{2}{c}{\small\textbf{Distinct}} & \multicolumn{2}{c}{\small\textbf{Entropy}} \\
&score&B-2&B-4&BP&Pre&Rec&F1&D-2&D-4&E-2&E-4\\
    \hline
    ml    & 0.04\% & 0.10\% & 0.00\% & 1.00  & 53.5\% & 57.0\% & 55.0\% & 0.12  & 0.58  & 5.90  & 9.17 \\
hi    & 0.04\% & 0.06\% & 0.00\% & 1.00  & 58.9\% & 60.0\% & 59.4\% & 0.13  & 0.63  & 6.00  & 8.83 \\
mk    & 0.05\% & 0.17\% & 0.00\% & 0.65  & 55.9\% & 56.6\% & 56.2\% & 0.04  & 0.28  & 4.32  & 7.08 \\
fa    & 0.01\% & 0.02\% & 0.01\% & 0.23  & 69.0\% & 65.0\% & 66.8\% & 0.18  & 0.51  & 5.97  & 8.09 \\
vi    & 0.14\% & 0.44\% & 0.20\% & 0.22  & 67.3\% & 63.9\% & 65.5\% & 0.08  & 0.34  & 5.48  & 7.58 \\
lv    & 0.13\% & 0.85\% & 0.01\% & 0.61  & 65.5\% & 62.1\% & 63.7\% & 0.14  & 0.47  & 6.31  & 8.56 \\
uk    & 0.23\% & 0.81\% & 0.02\% & 0.69  & 65.7\% & 62.6\% & 64.1\% & 0.15  & 0.43  & 6.34  & 8.55 \\
ko    & 0.19\% & 0.59\% & 0.01\% & 0.63  & 69.4\% & 66.2\% & 67.6\% & 0.22  & 0.54  & 7.11  & 8.78 \\
sq    & 0.58\% & 0.88\% & 0.26\% & 0.55  & 65.9\% & 63.0\% & 64.4\% & 0.16  & 0.45  & 6.35  & 8.61 \\
bg    & 0.40\% & 1.25\% & 0.03\% & 0.74  & 66.0\% & 63.8\% & 64.8\% & 0.20  & 0.49  & 6.72  & 8.79 \\
lt    & 0.12\% & 0.93\% & 0.01\% & 0.72  & 65.3\% & 62.4\% & 63.7\% & 0.15  & 0.41  & 6.22  & 8.41 \\
sk    & 0.37\% & 0.86\% & 0.05\% & 0.68  & 64.6\% & 61.9\% & 63.1\% & 0.19  & 0.53  & 6.80  & 8.92 \\
cs    & 0.32\% & 1.06\% & 0.01\% & 0.89  & 64.6\% & 63.0\% & 63.8\% & 0.13  & 0.39  & 6.24  & 8.38 \\
et    & 0.54\% & 1.38\% & 0.08\% & 0.70  & 65.9\% & 63.4\% & 64.5\% & 0.20  & 0.51  & 6.79  & 8.83 \\
ar    & 0.04\% & 0.11\% & 0.01\% & 0.59  & 67.3\% & 65.3\% & 66.2\% & 0.19  & 0.52  & 6.72  & 8.67 \\
sl    & 0.54\% & 1.12\% & 0.10\% & 0.70  & 64.9\% & 62.6\% & 63.7\% & 0.24  & 0.61  & 7.16  & 9.22 \\
he    & 0.21\% & 0.65\% & 0.01\% & 1.00  & 67.6\% & 66.3\% & 66.9\% & 0.14  & 0.35  & 6.20  & 7.91 \\
ja    & 1.74\% & 1.99\% & 0.62\% & 0.92  & 65.9\% & 64.1\% & 64.9\% & 0.47  & 0.66  & 8.29  & 9.00 \\
en    & 0.32\% & 0.95\% & 0.04\% & 0.60  & 85.5\% & 83.8\% & 84.6\% & 0.22  & 0.50  & 6.72  & 8.65 \\
no    & 0.77\% & 1.94\% & 0.11\% & 0.69  & 67.5\% & 64.8\% & 66.1\% & 0.19  & 0.48  & 6.72  & 8.85 \\
el    & 0.30\% & 0.60\% & 0.04\% & 0.70  & 66.0\% & 63.5\% & 64.6\% & 0.18  & 0.44  & 6.67  & 8.59 \\
zh    & 1.39\% & 2.69\% & 0.18\% & 0.95  & 58.6\% & 57.0\% & 57.7\% & 0.35  & 0.73  & 7.98  & 9.66 \\
da    & 1.39\% & 2.29\% & 0.38\% & 0.77  & 67.4\% & 65.0\% & 66.1\% & 0.19  & 0.51  & 6.83  & 8.96 \\
tl    & 0.19\% & 0.78\% & 0.01\% & 0.70  & 66.3\% & 64.2\% & 65.2\% & 0.19  & 0.49  & 6.46  & 8.56 \\
hu    & 0.47\% & 0.93\% & 0.03\% & 1.00  & 64.7\% & 63.2\% & 63.9\% & 0.19  & 0.46  & 6.49  & 8.72 \\
id    & 0.17\% & 0.61\% & 0.01\% & 0.71  & 66.5\% & 64.3\% & 65.4\% & 0.23  & 0.60  & 7.09  & 9.24 \\
fi    & 0.09\% & 0.48\% & 0.01\% & 0.57  & 66.3\% & 62.8\% & 64.5\% & 0.17  & 0.46  & 6.40  & 8.54 \\
pl    & 0.18\% & 0.67\% & 0.01\% & 0.75  & 65.4\% & 63.0\% & 64.1\% & 0.20  & 0.54  & 6.91  & 9.03 \\
ro    & 0.25\% & 0.71\% & 0.02\% & 0.71  & 64.4\% & 62.1\% & 63.2\% & 0.19  & 0.47  & 6.62  & 8.76 \\
sv    & 0.46\% & 1.28\% & 0.07\% & 0.57  & 67.0\% & 64.3\% & 65.6\% & 0.21  & 0.52  & 6.72  & 8.74 \\
hr    & 0.21\% & 0.70\% & 0.01\% & 0.68  & 65.1\% & 62.4\% & 63.7\% & 0.14  & 0.38  & 6.06  & 8.27 \\
nl    & 0.13\% & 0.90\% & 0.01\% & 0.42  & 67.6\% & 63.6\% & 65.4\% & 0.23  & 0.61  & 6.88  & 8.99 \\
it    & 0.19\% & 0.49\% & 0.01\% & 0.89  & 64.9\% & 63.6\% & 64.2\% & 0.17  & 0.43  & 6.43  & 8.45 \\
de    & 0.27\% & 0.37\% & 0.04\% & 0.81  & 65.7\% & 63.7\% & 64.6\% & 0.25  & 0.56  & 6.86  & 8.75 \\
ru    & 0.22\% & 0.62\% & 0.01\% & 0.91  & 65.6\% & 63.6\% & 64.5\% & 0.19  & 0.49  & 6.59  & 8.81 \\
pt    & 0.75\% & 1.19\% & 0.11\% & 0.93  & 66.7\% & 65.2\% & 65.9\% & 0.20  & 0.45  & 6.37  & 8.49 \\
fr    & 0.47\% & 0.92\% & 0.09\% & 0.65  & 67.0\% & 64.5\% & 65.7\% & 0.19  & 0.45  & 6.71  & 8.63 \\
tr    & 0.04\% & 0.13\% & 0.01\% & 0.62  & 41.8\% & 38.0\% & 39.6\% & 0.19  & 0.45  & 6.50  & 8.20 \\
es    & 0.57\% & 0.96\% & 0.06\% & 0.94  & 66.3\% & 64.7\% & 65.5\% & 0.21  & 0.60  & 7.20  & 9.44 \\

    \hline
    \end{tabular}%
    }
  \label{tab:lang_monot5}%
\end{table*}%

\begin{table*}[htbp]
  \centering
  \small
  \caption{Performance of Zero-Shot DialoGPT}
    \scalebox{1.1}
    {
    \begin{tabular}{lrrrrrrrrrrr}
\hline \multirow{2}{*}{\small\textbf{Language}} & \multicolumn{4}{c}{\small\textbf{sacreBLEU}} & \multicolumn{3}{c}{\small\textbf{Bert Score}} & \multicolumn{2}{c}{\small\textbf{Distinct}} & \multicolumn{2}{c}{\small\textbf{Entropy}} \\
&score&B-2&B-4&BP&Pre&Rec&F1&D-2&D-4&E-2&E-4\\
    \hline
mr    & 0.08\% & 0.04\% & 0.01\% & 1.00  & 60.3\% & 58.5\% & 59.3\% & 0.45  & 0.62  & 6.12  & 6.96 \\
kn    & 0.08\% & 0.04\% & 0.01\% & 1.00  & 60.3\% & 59.0\% & 59.6\% & 0.44  & 0.61  & 6.13  & 6.99 \\
ta    & 0.07\% & 0.04\% & 0.01\% & 1.00  & 59.8\% & 57.9\% & 58.8\% & 0.40  & 0.57  & 6.06  & 6.96 \\
bn    & 0.11\% & 0.10\% & 0.03\% & 1.00  & 60.3\% & 59.2\% & 59.7\% & 0.37  & 0.57  & 6.34  & 7.40 \\
ur    & 0.03\% & 0.02\% & 0.01\% & 1.00  & 59.8\% & 58.7\% & 59.2\% & 0.34  & 0.55  & 6.48  & 7.64 \\
te    & 0.10\% & 0.06\% & 0.02\% & 1.00  & 59.9\% & 58.1\% & 59.0\% & 0.32  & 0.53  & 6.49  & 7.75 \\
th    & 0.01\% & 0.01\% & 0.00\% & 1.00  & 59.9\% & 56.7\% & 58.2\% & 0.24  & 0.47  & 6.76  & 8.28 \\
ml    & 0.06\% & 0.03\% & 0.01\% & 1.00  & 59.5\% & 57.3\% & 58.3\% & 0.23  & 0.44  & 6.66  & 8.18 \\
hi    & 0.02\% & 0.01\% & 0.00\% & 0.95  & 60.7\% & 58.8\% & 59.7\% & 0.26  & 0.48  & 6.78  & 8.27 \\
mk    & 0.02\% & 0.01\% & 0.00\% & 0.82  & 60.9\% & 58.8\% & 59.8\% & 0.25  & 0.46  & 6.68  & 8.24 \\
fa    & 0.01\% & 0.01\% & 0.00\% & 1.00  & 59.2\% & 58.2\% & 58.7\% & 0.24  & 0.48  & 6.73  & 8.32 \\
vi    & 0.12\% & 0.12\% & 0.06\% & 0.61  & 61.5\% & 60.3\% & 60.9\% & 0.29  & 0.49  & 6.92  & 8.26 \\
lv    & 0.02\% & 0.04\% & 0.01\% & 0.26  & 62.9\% & 57.4\% & 60.0\% & 0.42  & 0.54  & 7.28  & 8.24 \\
uk    & 0.02\% & 0.01\% & 0.00\% & 0.84  & 61.2\% & 58.6\% & 59.8\% & 0.25  & 0.47  & 6.70  & 8.26 \\
ko    & 0.06\% & 0.03\% & 0.01\% & 1.00  & 60.9\% & 59.6\% & 60.2\% & 0.23  & 0.45  & 6.64  & 8.18 \\
sq    & 0.20\% & 0.58\% & 0.70\% & 0.16  & 62.9\% & 56.9\% & 59.7\% & 0.40  & 0.57  & 7.26  & 8.41 \\
bg    & 0.05\% & 0.03\% & 0.01\% & 0.79  & 61.5\% & 59.2\% & 60.3\% & 0.25  & 0.48  & 6.75  & 8.30 \\
lt    & 0.02\% & 0.10\% & 0.01\% & 0.26  & 63.3\% & 57.9\% & 60.4\% & 0.42  & 0.56  & 7.26  & 8.34 \\
sk    & 0.04\% & 0.13\% & 0.01\% & 0.29  & 63.5\% & 58.2\% & 60.7\% & 0.36  & 0.56  & 7.02  & 8.37 \\
cs    & 0.08\% & 0.22\% & 0.01\% & 0.41  & 63.0\% & 58.7\% & 60.7\% & 0.36  & 0.54  & 7.04  & 8.25 \\
et    & 0.12\% & 0.33\% & 0.17\% & 0.20  & 63.9\% & 58.6\% & 61.1\% & 0.44  & 0.57  & 7.36  & 8.29 \\
ar    & 0.03\% & 0.01\% & 0.00\% & 0.98  & 59.3\% & 57.4\% & 58.3\% & 0.25  & 0.48  & 6.74  & 8.34 \\
sl    & 0.03\% & 0.19\% & 0.01\% & 0.26  & 64.6\% & 59.2\% & 61.7\% & 0.28  & 0.53  & 6.50  & 8.35 \\
he    & 0.05\% & 0.02\% & 0.01\% & 1.00  & 61.0\% & 60.1\% & 60.5\% & 0.24  & 0.48  & 6.77  & 8.32 \\
ja    & 0.00\% & 0.01\% & 0.00\% & 0.54  & 60.2\% & 56.4\% & 58.2\% & 0.24  & 0.46  & 6.67  & 8.20 \\
en    & 0.38\% & 1.34\% & 0.10\% & 0.41  & 85.7\% & 83.6\% & 84.6\% & 0.36  & 0.64  & 7.33  & 8.96 \\
no    & 0.14\% & 0.81\% & 0.03\% & 0.33  & 65.1\% & 61.2\% & 63.1\% & 0.22  & 0.57  & 6.62  & 8.76 \\
el    & 0.07\% & 0.06\% & 0.01\% & 0.81  & 60.9\% & 57.2\% & 59.0\% & 0.23  & 0.44  & 6.58  & 8.16 \\
zh    & 0.00\% & 0.01\% & 0.00\% & 0.17  & 53.5\% & 49.9\% & 51.5\% & 0.24  & 0.45  & 6.67  & 8.19 \\
da    & 0.10\% & 0.93\% & 0.01\% & 0.39  & 65.5\% & 61.8\% & 63.5\% & 0.22  & 0.57  & 6.62  & 8.77 \\
tl    & 0.07\% & 0.35\% & 0.05\% & 0.21  & 63.4\% & 58.5\% & 60.8\% & 0.41  & 0.55  & 7.34  & 8.28 \\
hu    & 0.04\% & 0.12\% & 0.02\% & 0.27  & 62.7\% & 57.5\% & 59.9\% & 0.43  & 0.55  & 7.37  & 8.32 \\
id    & 0.04\% & 0.16\% & 0.01\% & 0.22  & 63.3\% & 59.0\% & 61.0\% & 0.44  & 0.56  & 7.47  & 8.34 \\
fi    & 0.03\% & 0.11\% & 0.01\% & 0.33  & 64.4\% & 59.9\% & 62.0\% & 0.36  & 0.54  & 7.17  & 8.32 \\
pl    & 0.03\% & 0.19\% & 0.01\% & 0.30  & 64.3\% & 59.8\% & 61.9\% & 0.40  & 0.60  & 7.24  & 8.48 \\
ro    & 0.04\% & 0.20\% & 0.01\% & 0.28  & 62.8\% & 58.4\% & 60.5\% & 0.26  & 0.50  & 6.53  & 8.22 \\
sv    & 0.04\% & 0.17\% & 0.01\% & 0.29  & 63.9\% & 60.0\% & 61.8\% & 0.20  & 0.42  & 6.17  & 7.83 \\
hr    & 0.12\% & 0.55\% & 0.04\% & 0.29  & 64.7\% & 59.5\% & 61.9\% & 0.25  & 0.51  & 6.35  & 8.24 \\
nl    & 0.09\% & 0.68\% & 0.15\% & 0.13  & 65.8\% & 60.5\% & 63.0\% & 0.32  & 0.52  & 6.78  & 8.11 \\
it    & 0.10\% & 0.39\% & 0.03\% & 0.37  & 64.6\% & 60.9\% & 62.6\% & 0.40  & 0.57  & 7.21  & 8.29 \\
de    & 0.09\% & 0.39\% & 0.01\% & 0.42  & 65.1\% & 61.6\% & 63.2\% & 0.35  & 0.56  & 7.04  & 8.35 \\
ru    & 0.06\% & 0.03\% & 0.01\% & 0.86  & 61.5\% & 59.2\% & 60.3\% & 0.24  & 0.46  & 6.72  & 8.23 \\
pt    & 0.18\% & 0.48\% & 0.03\% & 0.55  & 65.5\% & 62.6\% & 64.0\% & 0.24  & 0.45  & 6.27  & 7.94 \\
fr    & 0.07\% & 0.49\% & 0.01\% & 0.28  & 65.8\% & 61.5\% & 63.5\% & 0.20  & 0.43  & 6.25  & 8.09 \\
tr    & 0.05\% & 0.06\% & 0.01\% & 0.54  & 36.8\% & 34.5\% & 35.4\% & 0.37  & 0.51  & 7.08  & 8.08 \\
es    & 0.13\% & 0.63\% & 0.01\% & 0.53  & 65.6\% & 62.3\% & 63.8\% & 0.25  & 0.54  & 6.82  & 8.50 \\

    \hline
    \end{tabular}%
    }
  \label{tab:lang_zerodialogpt}%
\end{table*}%

\begin{table*}[htbp]
  \centering
  \small
  \caption{Performance of Fine-tuned DialoGPT}
    \scalebox{1.1}
    {
    \begin{tabular}{lrrrrrrrrrrr}
\hline \multirow{2}{*}{\small\textbf{Language}} & \multicolumn{4}{c}{\small\textbf{sacreBLEU}} & \multicolumn{3}{c}{\small\textbf{Bert Score}} & \multicolumn{2}{c}{\small\textbf{Distinct}} & \multicolumn{2}{c}{\small\textbf{Entropy}} \\
&score&B-2&B-4&BP&Pre&Rec&F1&D-2&D-4&E-2&E-4\\
    \hline
    ml    & 0.10\% & 0.08\% & 0.04\% & 0.81  & 66.8\% & 65.8\% & 66.2\% & 0.50  & 0.64  & 7.56  & 8.61 \\
hi    & 0.49\% & 0.54\% & 0.10\% & 1.00  & 66.7\% & 66.6\% & 66.6\% & 0.33  & 0.68  & 7.47  & 9.16 \\
mk    & 0.18\% & 0.39\% & 0.01\% & 1.00  & 63.2\% & 63.4\% & 63.3\% & 0.36  & 0.73  & 7.69  & 9.38 \\
fa    & 0.06\% & 0.06\% & 0.01\% & 1.00  & 68.0\% & 67.6\% & 67.8\% & 0.28  & 0.66  & 7.15  & 9.00 \\
vi    & 0.71\% & 1.05\% & 0.29\% & 0.73  & 67.7\% & 66.2\% & 66.9\% & 0.20  & 0.52  & 6.91  & 8.56 \\
lv    & 0.25\% & 0.71\% & 0.01\% & 0.94  & 64.1\% & 62.9\% & 63.4\% & 0.28  & 0.64  & 7.40  & 9.23 \\
uk    & 0.14\% & 0.42\% & 0.01\% & 1.00  & 64.1\% & 63.7\% & 63.9\% & 0.36  & 0.73  & 7.84  & 9.40 \\
ko    & 0.18\% & 0.38\% & 0.01\% & 0.93  & 67.6\% & 67.3\% & 67.4\% & 0.34  & 0.66  & 7.54  & 9.04 \\
sq    & 0.93\% & 0.76\% & 0.28\% & 0.94  & 64.4\% & 63.3\% & 63.8\% & 0.26  & 0.59  & 7.49  & 9.31 \\
bg    & 0.46\% & 0.70\% & 0.05\% & 1.00  & 64.8\% & 64.6\% & 64.7\% & 0.31  & 0.65  & 7.51  & 9.14 \\
lt    & 0.24\% & 0.89\% & 0.01\% & 0.90  & 64.8\% & 63.5\% & 64.1\% & 0.33  & 0.68  & 7.75  & 9.37 \\
sk    & 0.36\% & 0.82\% & 0.06\% & 0.73  & 64.4\% & 62.4\% & 63.3\% & 0.32  & 0.67  & 7.67  & 9.28 \\
cs    & 0.23\% & 0.80\% & 0.01\% & 0.90  & 64.7\% & 63.5\% & 64.0\% & 0.26  & 0.60  & 7.26  & 9.07 \\
et    & 0.31\% & 1.20\% & 0.01\% & 0.86  & 65.7\% & 64.0\% & 64.8\% & 0.36  & 0.69  & 7.81  & 9.35 \\
ar    & 0.18\% & 0.16\% & 0.03\% & 1.00  & 66.4\% & 66.4\% & 66.3\% & 0.39  & 0.72  & 8.02  & 9.31 \\
sl    & 0.42\% & 1.02\% & 0.04\% & 0.82  & 64.4\% & 62.8\% & 63.5\% & 0.33  & 0.68  & 7.80  & 9.39 \\
he    & 0.38\% & 0.45\% & 0.06\% & 1.00  & 66.3\% & 66.7\% & 66.4\% & 0.27  & 0.57  & 7.26  & 8.79 \\
ja    & 2.14\% & 2.08\% & 0.89\% & 0.97  & 64.6\% & 63.8\% & 64.1\% & 0.50  & 0.65  & 8.06  & 8.84 \\
en    & 0.48\% & 1.14\% & 0.04\% & 0.82  & 85.2\% & 83.9\% & 84.5\% & 0.33  & 0.65  & 7.58  & 9.27 \\
no    & 1.23\% & 1.70\% & 0.24\% & 0.95  & 66.2\% & 64.7\% & 65.4\% & 0.27  & 0.63  & 7.62  & 9.48 \\
el    & 0.28\% & 0.33\% & 0.04\% & 1.00  & 65.1\% & 64.5\% & 64.8\% & 0.27  & 0.60  & 7.33  & 9.03 \\
zh    & 1.40\% & 2.08\% & 0.23\% & 1.00  & 57.6\% & 57.1\% & 57.3\% & 0.49  & 0.77  & 8.39  & 9.48 \\
da    & 1.54\% & 1.79\% & 0.36\% & 1.00  & 66.3\% & 64.8\% & 65.5\% & 0.24  & 0.62  & 7.44  & 9.40 \\
tl    & 0.52\% & 0.87\% & 0.10\% & 0.73  & 67.1\% & 65.2\% & 66.1\% & 0.30  & 0.66  & 7.48  & 9.24 \\
hu    & 0.35\% & 0.97\% & 0.02\% & 1.00  & 64.4\% & 63.5\% & 63.9\% & 0.30  & 0.62  & 7.44  & 9.29 \\
id    & 0.13\% & 0.47\% & 0.00\% & 0.93  & 65.8\% & 64.8\% & 65.3\% & 0.36  & 0.76  & 8.09  & 9.65 \\
fi    & 0.27\% & 0.48\% & 0.04\% & 0.74  & 65.4\% & 63.4\% & 64.3\% & 0.38  & 0.71  & 7.86  & 9.41 \\
pl    & 0.40\% & 0.73\% & 0.04\% & 0.85  & 65.0\% & 63.4\% & 64.1\% & 0.33  & 0.69  & 7.74  & 9.39 \\
ro    & 0.26\% & 0.70\% & 0.01\% & 0.93  & 64.0\% & 62.6\% & 63.3\% & 0.27  & 0.65  & 7.62  & 9.49 \\
sv    & 0.48\% & 1.07\% & 0.04\% & 0.90  & 65.9\% & 64.2\% & 65.0\% & 0.27  & 0.63  & 7.43  & 9.26 \\
hr    & 0.12\% & 0.53\% & 0.00\% & 0.87  & 64.5\% & 62.9\% & 63.6\% & 0.30  & 0.70  & 7.73  & 9.58 \\
nl    & 0.69\% & 0.91\% & 0.11\% & 0.96  & 65.7\% & 64.5\% & 65.0\% & 0.28  & 0.68  & 7.62  & 9.50 \\
it    & 0.42\% & 0.56\% & 0.06\% & 0.99  & 64.8\% & 63.9\% & 64.3\% & 0.27  & 0.59  & 7.45  & 9.06 \\
de    & 0.38\% & 0.53\% & 0.03\% & 1.00  & 64.9\% & 63.9\% & 64.3\% & 0.35  & 0.67  & 7.64  & 9.17 \\
ru    & 0.10\% & 0.23\% & 0.01\% & 1.00  & 64.4\% & 64.1\% & 64.2\% & 0.40  & 0.75  & 7.98  & 9.40 \\
pt    & 0.68\% & 0.99\% & 0.09\% & 1.00  & 66.3\% & 65.5\% & 65.8\% & 0.27  & 0.57  & 7.14  & 9.02 \\
fr    & 1.00\% & 1.39\% & 0.24\% & 0.86  & 66.1\% & 64.8\% & 65.4\% & 0.25  & 0.56  & 7.39  & 9.14 \\
tr    & 0.09\% & 0.20\% & 0.01\% & 0.66  & 41.4\% & 38.8\% & 39.9\% & 0.42  & 0.65  & 7.85  & 8.91 \\
es    & 0.62\% & 1.13\% & 0.07\% & 0.94  & 66.1\% & 64.7\% & 65.3\% & 0.29  & 0.68  & 7.70  & 9.49 \\

    \hline
    \end{tabular}%
    }
  \label{tab:lang_ftdialogpt}%
\end{table*}%

\begin{table*}[htbp]
  \centering
  \small
  \caption{Performance of Zero-Shot DialoGPT+MarianMT}
    \scalebox{1.1}
    {
    \begin{tabular}{lrrrrrrrrrrr}
\hline \multirow{2}{*}{\small\textbf{Language}} & \multicolumn{4}{c}{\small\textbf{sacreBLEU}} & \multicolumn{3}{c}{\small\textbf{Bert Score}} & \multicolumn{2}{c}{\small\textbf{Distinct}} & \multicolumn{2}{c}{\small\textbf{Entropy}} \\
&score&B-2&B-4&BP&Pre&Rec&F1&D-2&D-4&E-2&E-4\\
    \hline
    mr    & 0.17\% & 0.30\% & 0.04\% & 0.65  & 66.1\% & 63.8\% & 64.9\% & 0.50  & 0.66  & 6.50  & 7.20 \\
ur    & 0.16\% & 0.26\% & 0.02\% & 0.99  & 65.9\% & 64.5\% & 65.1\% & 0.35  & 0.65  & 6.95  & 8.23 \\
ml    & 0.08\% & 0.15\% & 0.01\% & 0.50  & 65.6\% & 62.9\% & 64.2\% & 0.44  & 0.66  & 7.72  & 8.98 \\
hi    & 0.19\% & 0.49\% & 0.01\% & 0.81  & 67.0\% & 65.1\% & 66.0\% & 0.27  & 0.55  & 7.08  & 8.78 \\
mk    & 0.09\% & 0.41\% & 0.01\% & 0.50  & 66.1\% & 63.1\% & 64.5\% & 0.37  & 0.60  & 7.30  & 8.56 \\
vi    & 0.05\% & 0.19\% & 0.00\% & 1.00  & 66.2\% & 64.4\% & 65.2\% & 0.15  & 0.40  & 6.49  & 7.79 \\
uk    & 0.08\% & 0.66\% & 0.01\% & 0.43  & 65.9\% & 62.2\% & 63.9\% & 0.43  & 0.67  & 7.80  & 9.07 \\
sq    & 0.07\% & 0.32\% & 0.01\% & 0.36  & 65.7\% & 61.7\% & 63.5\% & 0.33  & 0.62  & 7.39  & 8.99 \\
bg    & 0.19\% & 1.04\% & 0.01\% & 0.46  & 66.4\% & 63.0\% & 64.6\% & 0.35  & 0.60  & 7.40  & 8.75 \\
sk    & 0.11\% & 0.67\% & 0.01\% & 0.40  & 65.2\% & 61.7\% & 63.4\% & 0.41  & 0.67  & 7.70  & 9.06 \\
cs    & 0.22\% & 0.94\% & 0.02\% & 0.54  & 65.4\% & 62.7\% & 64.0\% & 0.39  & 0.65  & 7.60  & 9.01 \\
et    & 0.19\% & 1.07\% & 0.02\% & 0.45  & 66.7\% & 63.2\% & 64.8\% & 0.42  & 0.65  & 7.66  & 8.96 \\
ar    & 0.03\% & 0.04\% & 0.00\% & 0.73  & 66.1\% & 64.0\% & 65.0\% & 0.35  & 0.58  & 7.56  & 8.61 \\
ja    & 0.07\% & 0.22\% & 0.00\% & 1.00  & 60.7\% & 61.0\% & 60.8\% & 0.06  & 0.22  & 5.42  & 7.67 \\
zh    & 0.34\% & 2.18\% & 0.12\% & 0.28  & 57.8\% & 53.6\% & 55.5\% & 0.53  & 0.66  & 7.94  & 8.70 \\
da    & 0.28\% & 1.46\% & 0.09\% & 0.31  & 68.5\% & 64.0\% & 66.1\% & 0.39  & 0.67  & 7.55  & 9.02 \\
tl    & 0.06\% & 0.25\% & 0.01\% & 0.39  & 66.2\% & 63.4\% & 64.7\% & 0.36  & 0.65  & 7.59  & 9.04 \\
hu    & 0.12\% & 0.90\% & 0.01\% & 0.45  & 66.2\% & 62.8\% & 64.4\% & 0.41  & 0.65  & 7.56  & 8.95 \\
id    & 0.05\% & 0.19\% & 0.01\% & 0.36  & 66.5\% & 62.3\% & 64.3\% & 0.38  & 0.65  & 7.52  & 8.89 \\
fi    & 0.10\% & 0.64\% & 0.01\% & 0.40  & 66.9\% & 62.9\% & 64.7\% & 0.44  & 0.64  & 7.71  & 8.88 \\
sv    & 0.15\% & 0.83\% & 0.01\% & 0.40  & 67.9\% & 64.0\% & 65.8\% & 0.38  & 0.67  & 7.57  & 9.02 \\
nl    & 0.45\% & 1.09\% & 0.20\% & 0.43  & 67.5\% & 63.9\% & 65.6\% & 0.38  & 0.65  & 7.63  & 8.96 \\
it    & 0.12\% & 0.38\% & 0.01\% & 0.60  & 65.6\% & 63.4\% & 64.4\% & 0.38  & 0.65  & 7.53  & 8.93 \\
de    & 0.19\% & 0.49\% & 0.01\% & 0.85  & 65.6\% & 63.8\% & 64.6\% & 0.41  & 0.67  & 7.70  & 9.07 \\
ru    & 0.13\% & 0.54\% & 0.01\% & 0.50  & 66.0\% & 63.0\% & 64.4\% & 0.40  & 0.65  & 7.54  & 8.98 \\
fr    & 0.28\% & 1.16\% & 0.05\% & 0.43  & 67.3\% & 64.2\% & 65.6\% & 0.33  & 0.61  & 7.41  & 8.95 \\
es    & 0.27\% & 0.98\% & 0.08\% & 0.38  & 67.5\% & 63.9\% & 65.6\% & 0.40  & 0.66  & 7.65  & 8.91 \\

    \hline
    \end{tabular}%
    }
  \label{tab:lang_mt_dialogpt}%
\end{table*}%

\begin{table*}[htbp]
  \centering
  \small
  \caption{Performance of Fine-tuned DialoGPT+MarianMT}
    \scalebox{1.1}
    {
    \begin{tabular}{lrrrrrrrrrrr}
\hline \multirow{2}{*}{\small\textbf{Language}} & \multicolumn{4}{c}{\small\textbf{sacreBLEU}} & \multicolumn{3}{c}{\small\textbf{Bert Score}} & \multicolumn{2}{c}{\small\textbf{Distinct}} & \multicolumn{2}{c}{\small\textbf{Entropy}} \\
&score&B-2&B-4&BP&Pre&Rec&F1&D-2&D-4&E-2&E-4\\
    \hline
   ml    & 0.13\% & 0.20\% & 0.01\% & 0.82  & 65.5\% & 62.3\% & 63.8\% & 0.40  & 0.57  & 7.60  & 8.64 \\
hi    & 0.08\% & 0.22\% & 0.00\% & 1.00  & 66.8\% & 65.2\% & 65.9\% & 0.15  & 0.31  & 6.32  & 7.58 \\
mk    & 0.16\% & 0.34\% & 0.00\% & 1.00  & 64.7\% & 62.8\% & 63.7\% & 0.23  & 0.45  & 7.16  & 8.51 \\
vi    & 0.10\% & 0.30\% & 0.00\% & 1.00  & 66.6\% & 64.5\% & 65.5\% & 0.13  & 0.30  & 6.40  & 7.72 \\
uk    & 0.11\% & 0.44\% & 0.00\% & 1.00  & 63.0\% & 62.3\% & 62.6\% & 0.23  & 0.45  & 7.49  & 8.97 \\
sq    & 0.04\% & 0.14\% & 0.00\% & 1.00  & 63.9\% & 62.0\% & 62.8\% & 0.20  & 0.44  & 7.05  & 8.56 \\
bg    & 0.26\% & 0.67\% & 0.01\% & 1.00  & 64.2\% & 63.5\% & 63.7\% & 0.20  & 0.43  & 7.01  & 8.57 \\
sk    & 0.19\% & 0.73\% & 0.00\% & 0.88  & 64.5\% & 62.6\% & 63.5\% & 0.31  & 0.58  & 7.61  & 9.07 \\
cs    & 0.42\% & 0.93\% & 0.04\% & 0.90  & 65.2\% & 63.4\% & 64.2\% & 0.30  & 0.57  & 7.36  & 8.84 \\
et    & 0.36\% & 0.87\% & 0.03\% & 0.87  & 65.8\% & 63.7\% & 64.7\% & 0.37  & 0.63  & 7.89  & 9.10 \\
ar    & 0.05\% & 0.04\% & 0.01\% & 0.89  & 65.8\% & 63.9\% & 64.8\% & 0.22  & 0.37  & 6.95  & 7.79 \\
ja    & 0.05\% & 0.16\% & 0.00\% & 1.00  & 59.9\% & 60.8\% & 60.3\% & 0.04  & 0.15  & 5.28  & 7.25 \\
zh    & 0.53\% & 1.41\% & 0.07\% & 0.72  & 56.9\% & 53.6\% & 55.1\% & 0.33  & 0.46  & 7.62  & 8.27 \\
da    & 0.50\% & 1.43\% & 0.04\% & 0.75  & 67.5\% & 64.8\% & 66.0\% & 0.31  & 0.63  & 7.60  & 9.16 \\
tl    & 0.09\% & 0.26\% & 0.00\% & 0.71  & 65.7\% & 63.6\% & 64.6\% & 0.25  & 0.52  & 7.20  & 8.63 \\
hu    & 0.33\% & 1.05\% & 0.02\% & 0.81  & 65.4\% & 63.0\% & 64.1\% & 0.32  & 0.58  & 7.41  & 8.95 \\
id    & 0.11\% & 0.22\% & 0.00\% & 0.98  & 64.5\% & 61.7\% & 63.0\% & 0.21  & 0.41  & 6.84  & 8.00 \\
fi    & 0.15\% & 0.51\% & 0.00\% & 1.00  & 65.7\% & 63.5\% & 64.5\% & 0.27  & 0.45  & 6.88  & 7.88 \\
sv    & 0.58\% & 0.99\% & 0.10\% & 0.82  & 67.5\% & 64.6\% & 66.0\% & 0.30  & 0.58  & 7.35  & 8.71 \\
nl    & 0.33\% & 0.74\% & 0.03\% & 0.79  & 66.9\% & 64.5\% & 65.6\% & 0.31  & 0.62  & 7.63  & 9.04 \\
it    & 0.28\% & 0.53\% & 0.02\% & 0.94  & 65.1\% & 63.7\% & 64.3\% & 0.28  & 0.54  & 7.25  & 8.64 \\
de    & 0.17\% & 0.37\% & 0.00\% & 1.00  & 65.3\% & 63.8\% & 64.5\% & 0.32  & 0.54  & 7.23  & 8.21 \\
ru    & 0.08\% & 0.24\% & 0.00\% & 1.00  & 64.2\% & 63.0\% & 63.5\% & 0.17  & 0.32  & 6.65  & 7.82 \\
fr    & 0.42\% & 0.92\% & 0.03\% & 0.88  & 66.6\% & 64.8\% & 65.6\% & 0.26  & 0.52  & 7.32  & 8.82 \\
es    & 0.49\% & 1.07\% & 0.07\% & 0.73  & 67.1\% & 64.7\% & 65.8\% & 0.33  & 0.65  & 7.69  & 9.20 \\

    \hline
    \end{tabular}%
    }
  \label{tab:lang_ftmt_dialogpt}%
\end{table*}%

\end{document}